\documentclass{article}

\PassOptionsToPackage{numbers, compress, sort}{natbib}

\usepackage[final]{neurips_data_2023}

\usepackage[utf8]{inputenc} %
\usepackage[T1]{fontenc}    %
\usepackage[hypertexnames=false,colorlinks=true,citecolor=NavyBlue,backref=page]{hyperref}       %
\usepackage{url}            %
\usepackage{booktabs}       %
\usepackage{amsfonts}       %
\usepackage{nicefrac}       %
\usepackage{microtype}      %
\usepackage[table,xcdraw,dvipsnames]{xcolor}         %
\usepackage{paralist}
\newcommand{\cmark}{\ding{51}}%
\newcommand{\xmark}{\ding{55}}%
\usepackage{makecell}
\usepackage{setspace}
\usepackage{multirow}
\usepackage{graphicx}
\usepackage{booktabs}
\usepackage{subcaption}
\usepackage{pifont}%

\newcommand{\dsquestion}[1]{
    \item \noindent \textcolor{black}{\textbf{#1}} \smallskip
}
\newcommand{\dsquestionex}[2]{
    \item \noindent \textcolor{black}{\textbf{#1} \textit{#2}} \smallskip
}
\newcommand{\dsanswer}[1]{
    \hspace{-14pt} -- \hspace{10pt}
    \textcolor{blue}{#1} \medskip
}
\newcommand{\qref}[1]{\textcolor{red}{Q}\ref{#1}}

\newcommand{\dataset}{VidChapters-7M}
\newcommand{\dsetsize}{817K}
\newcommand{\chaptersize}{7M}
\definecolor{greencode}{RGB}{13, 144, 79}

\def\sepappendix{0}
\title{\dataset{}: 
Video Chapters at Scale}

\author{
Antoine Yang$^\dag$,~
Arsha Nagrani$^\S$,~
Ivan Laptev$^\dag$,~
Josef Sivic$^\P$,~
Cordelia Schmid$^\dag$ \\
\small{$^\dag$Inria Paris, DI ENS, CNRS, PSL Research University}\quad 
\small{$^\S$ VGG, University of Oxford}\\ 
\small{$^\P$Czech Institute of Informatics, Robotics and Cybernetics at the Czech Technical University in Prague}\\
\small{\url{https://antoyang.github.io/vidchapters.html}}
}

\begin{document}

\maketitle

\begin{abstract}
Segmenting long videos into chapters enables users to quickly navigate to the information of their interest.
This important topic has been understudied due to the lack of publicly released datasets.
To address this issue, we present \dataset{}, a dataset of \dsetsize{} user-chaptered videos including \chaptersize{} chapters in total.
\dataset{} is automatically created from videos online in a scalable manner by scraping user-annotated chapters and hence without any additional manual annotation.
We introduce the following three tasks based on this data. 
First, the video chapter generation task consists of temporally segmenting the video and generating a chapter title for each segment.
To further dissect the problem, we also define two variants of this task: video chapter generation given ground-truth boundaries, which requires generating a chapter title given an annotated video segment, and video chapter grounding, which requires temporally localizing a chapter given its annotated title.
We benchmark both simple baselines and state-of-the-art video-language models for these three tasks.
We also show that pretraining on \dataset{} transfers well to dense video captioning tasks in both zero-shot and finetuning settings, largely improving the state of the art on the YouCook2 and ViTT benchmarks. 
Finally, our experiments reveal that downstream performance scales well with the size of the pretraining dataset.
Our dataset, code, and models are publicly available at \url{https://antoyang.github.io/vidchapters.html}.

\end{abstract}

\begin{figure*}[h]
\centering
\vspace{-0.5cm}
\includegraphics[width=1.\linewidth]{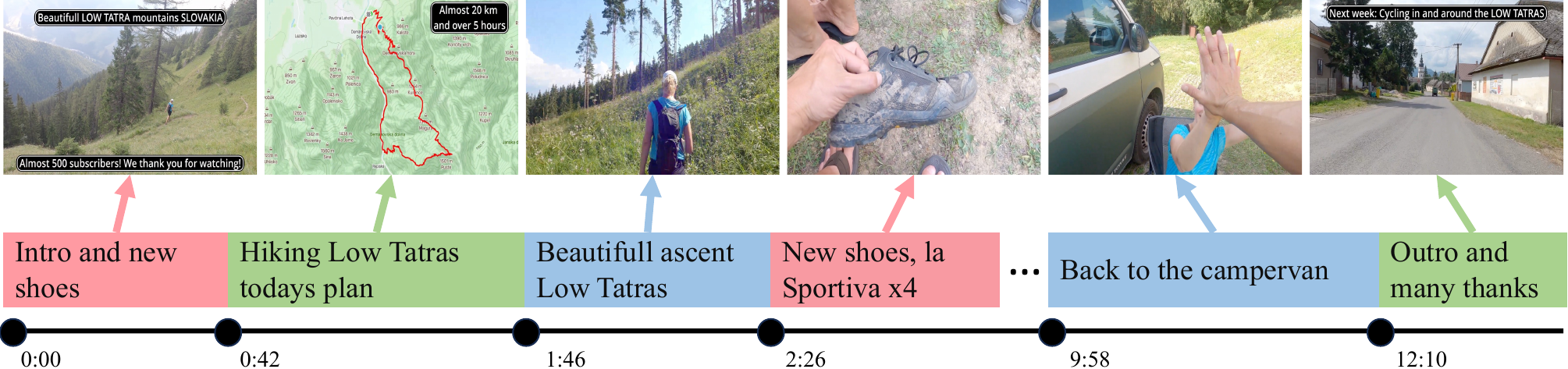}
\vspace{-0.5cm}
\caption{
\textbf{A video with user-annotated chapters in \dataset{}:} the video is temporally segmented into chapters, which are annotated with a chapter title in free-form natural language.
}
\vspace{-0.5cm}
\label{fig:teaser}
\end{figure*}

\section{Introduction}\label{sec:intro}
As online media consumption grows, the volume of video content available is increasing rapidly. 
While searching for specific videos is already a challenging problem, searching within a long video is an even \textit{less} explored task.
Manual navigation can often be time consuming, particularly for long videos. 
A compelling solution for organizing content online is to segment long videos into \textit{chapters} (see Figure~\ref{fig:teaser}). 
Chapters are contiguous, non-overlapping segments, completely partitioning a video. 
Each chapter is also labeled with a short description of the chapter content, enabling users to quickly navigate to areas of interest and easily replay different parts of a video.
Chapters also give \textit{structure} to a video, which is useful for long videos that contain inherently listed content, such as listicles~\cite{vijgen2014listicle}, instructional videos~\cite{miech19howto100m}, music compilations and so on.

\begin{figure*}[t]
\centering
\includegraphics[width=1.\linewidth]{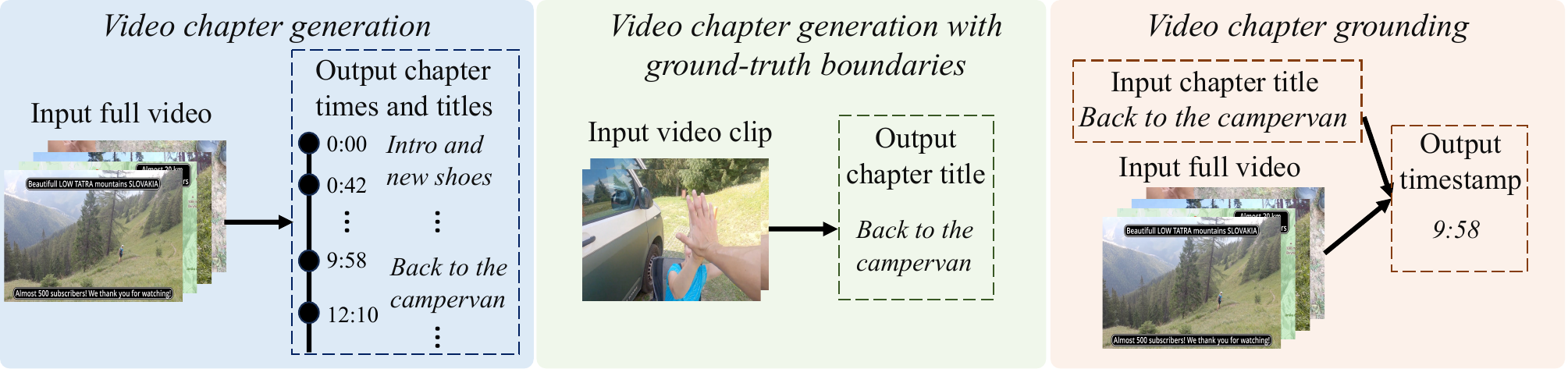}
\vspace{-0.5cm}
\caption{
\textbf{Illustration of the three tasks defined for \dataset{}.}
}
\vspace{-0.5cm}
\label{fig:teaser2}
\end{figure*}

Given the plethora of content already online, our goal is to explore automatic solutions related to video chaptering - generating chapters automatically, and grounding chapter titles temporally in long videos. 
While the benefits of automatically chaptering videos are obvious, data for this task is scarce. 
Video captioning datasets (such as WebVid-10M~\cite{bain2021frozen} and VideoCC~\cite{nagrani2022learning}) consist of short videos (10s in length), and hence are unsuitable. 
Web datasets consisting of longer videos (HowTo100M~\cite{miech19howto100m}, YT-Temporal-1B~\cite{zellers2022merlot}) come with aligned speech transcripts (ASR), which are only weakly related to visual content, and if used as chapter titles would tend to over-segment videos. 
Moment retrieval~\cite{gao2017tall, hendricks17localizing} or dense video captioning~\cite{krishna2017dense, zhou18towards} datasets are perhaps the most useful, but do not focus on creating explicit \textit{structure}, and instead describe low-level actions comprehensively. 
Such datasets are also manually annotated, and hence not scalable and small in size (see Table~\ref{table:datasetcomparison}).

To remedy this, we curate \dataset{}, a large-scale dataset of user-annotated video chapters automatically scraped from the Web. 
Our dataset consists of \chaptersize{} chapters for over \dsetsize{} videos. 
Compared to existing datasets, videos in \dataset{} are long (23 minutes on average) and contain rich chapter annotations consisting of a starting timestamp and a title per chapter. 
Our dataset is also diverse, with 12 different video categories having at least 20K videos each, 
which itself is the size of existing dense video captioning datasets~\cite{Ego4D2022CVPR, huang2020multimodal, krishna2017dense, zhou18towards}. 
On top of this dataset we also define 3 video tasks (see Figure~\ref{fig:teaser2}): 
(i) \textit{video chapter generation}
which requires temporally segmenting the video and generating a chapter title for each segment; 
(ii) \textit{video chapter generation given ground-truth boundaries}
, which requires generating a chapter title given an annotated video segment; 
and (iii) \textit{video chapter grounding}
, which requires temporally localizing a chapter given the chapter title. 
All three tasks involve parsing and understanding \textit{long} videos,
and multi-modal reasoning (video and text), and hence are valuable steps towards story understanding.

For all three tasks, we implement simple baselines as well as recent, state-of-the-art video-text methods~\cite{lei2021detecting, wang2021end, yang2023vid2seq}.
We find that the tasks are far from being solved, demonstrating the value of this problem.
Interestingly, we also show that our video chapter generation models trained on \dataset{} transfer well to dense video captioning tasks in both zero-shot and finetuning settings, largely improving the state of the art on the YouCook2~\cite{zhou18towards} and ViTT benchmarks~\cite{huang2020multimodal}.
Moreover, we show that pretraining using both speech transcripts and chapter annotations significantly outperforms the widely used pretraining method based only on speech transcripts~\cite{miech20endtoend, yang2023vid2seq, zellers2022merlot}.
This demonstrates the additional value of our dataset as a generic video-language \textit{pretraining} set. 
Interestingly, we also find that the transfer performance scales with the size of the chapter dataset.

In summary, our contributions are:
\begin{itemize}
\item[\textit{(i)}] We present \dataset{}, a large-scale dataset of user-annotated video chapters obtained from the Web consisting of \dsetsize{} videos and \chaptersize{} chapters;
\item[\textit{(ii)}] Based on this dataset, we evaluate a range of simple baselines and state-of-the-art video-language models on the tasks of video chapter generation with and without ground-truth boundaries, and video chapter grounding;
\item[\textit{(iii)}] We show that video chapter generation models trained on \dataset{} transfer well to dense video captioning tasks in both zero-shot and finetuning settings, largely improving the state of the art on the YouCook2~\cite{zhou18towards} and ViTT benchmarks~\cite{huang2020multimodal}, outperforming prior pretraining methods based on narrated videos~\cite{yang2023vid2seq}, and showing promising scaling behavior.
\end{itemize}
Our dataset, code and models are publicly available on our website~\cite{chapterswebpage}.

\section{Related Work}\label{sec:background}
\begin{table}
\begin{center}
\resizebox{.8\linewidth}{!}{
\begin{tabular}{l|c|c|c|c}
Dataset 
& \makecell{ \small{Number of} \\ \small{videos} } 
& \makecell{ \small{Video} \\ \small{duration (min)} } 
& \makecell{ \small{Number of} \\ \small{descriptions} } 
& Annotations \\
\hline
HowTo100M~\cite{miech19howto100m} & 1M & 7 & 136M & Speech transcripts \\ %
YT-Temporal-1B~\cite{zellers2022merlot} & \textbf{19M} & 6 & \textbf{$\sim$ 900M} & Speech transcripts \\ %
HD-VILA-100M~\cite{xue2022advancing} & 3M & 7 & 103M & Speech transcripts \\ %
\hline
ActivityNet Captions~\cite{krishna2017dense} & 20K & 3 & 100K & Dense Captions \\ %
YouCook2~\cite{zhou18towards} & 2K & 6 & 15K & Dense Captions \\ %
ViTT~\cite{huang2020multimodal} & 8K & 4 & 56K & Dense Captions \\ %
Ego4D~\cite{Ego4D2022CVPR} & 10K & \textbf{23} & 4M & Dense Captions \\ %
\hline
\dataset{} (Ours) & \dsetsize{} & \textbf{23} & \chaptersize{} & \textbf{\makecell{Speech transcripts + \\ User-annotated Chapters}} \\
\end{tabular}
}
\end{center}
\caption{\textbf{Comparison of \dataset{} with existing datasets}. 
We consider open-sourced video datasets that contain dense natural language descriptions aligned over time. 
\dataset{} is much larger than current dense video captioning datasets. 
Compared to datasets with ASR (top 3 rows), it is smaller in the total number of videos but contains longer videos with richer annotations (chapters).}
\label{table:datasetcomparison}
\vspace{-0.5cm}
\end{table}

\noindent \textbf{Large-scale vision-language datasets.}
The development of powerful multi-modal models~\cite{alayrac2022flamingo, chen2019uniter, gan2020large, hu2022scaling, huang2021seeing, jia2021scaling, lei2021less, li2019unicodervl, li2021align, li2022blip, li2020oscar, lu2019vilbert, lu202012, radford2021learning, singh2022flava, su2019vl, tan2019lxmert, tsimpoukelli2021multimodal, wang2021simvlm, wang2022git, yu2020ernie, yuan2021florence, zhou2020unified} has been made possible by pretraining on large-scale image-caption datasets scraped from the Web such as SBU~\cite{ordonez2011im2text}, Conceptual Captions~\cite{sharma2018conceptual}, Conceptual-12M~\cite{changpinyo2021conceptual}, LAIT~\cite{qi2020imagebert}, Wikipedia-ImageText~\cite{srinivasan2021wit},  RedCaps~\cite{desai2021redcaps} and LAION-5B~\cite{schuhmannlaion}.
Similarly, many strong video-language models~\cite{akbari2021vatt, ge2022bridging, han2022temporal, ko2022video, lei2021detecting, li2021align2, li2020hero, li2022lavender, lin2022egocentric, miech20endtoend, seo2020look, seo2022end, sun2019videobert, sun2022long, tang2022tvlt, wang2022all, wang2022object, xu2021videoclip, yang2021just, Yang2022LearningTA, yang2022frozenbilm, zhao2022learning} have been pretrained on Web-scraped video-text datasets.
These datasets are largely composed of short videos paired with captions, e.g.~WebVid-10M~\cite{bain2021frozen} and VideoCC~\cite{nagrani2022learning}, or narrated videos with speech transcripts aligned over time (ASR), e.g.~HowTo100M~\cite{miech19howto100m}, YT-Temporal-1B~\cite{zellers2021merlot, zellers2022merlot} and HD-VILA-100M~\cite{xue2022advancing}.
Our proposed~\dataset{} dataset is also downloaded from the Web, via a scalable pipeline without the need for expensive manual annotation.
Unlike these datasets, \dataset{} consists of long videos with user-annotated chapters aligned over time (see Table~\ref{table:datasetcomparison}), which significantly differ from ASR (see Section~\ref{sec:analysis}).
Furthermore, most videos in \dataset{} \emph{also} contain ASR.
Finally, \dataset{} is also related to the recent ChapterGen dataset~\cite{cao2022multi}, which also consists of user-annotated chapters.
However, ChapterGen is several orders of magnitude smaller than \dataset{} (10K vs \dsetsize{} videos) and is not open-sourced at the time of writing.

\noindent \textbf{Video tasks.}
The video chapter generation task requires temporally segmenting the video into chapters, hence is related to video shot detection~\cite{rasheed2003scene, rui1998exploring, sidiropoulos2011temporal}, movie scene segmentation~\cite{chen2021shot, rao2020local}, temporal action localization~\cite{chao2018rethinking, cheng2022tallformer, liu2022end, shou2016temporal, zhang2022actionformer, zeng2019graph} and temporal action segmentation~\cite{behrmann2022unified, farha2019ms, gao2021global, lea2017temporal, li2021temporal, wang2020boundary}.
However, unlike these tasks, video chapter generation also requires generating a free-form natural language chapter title for each segment.
Hence this task is also related to video captioning~\cite{gao2017video, lin2022swinbert, luo2020univilm, pan2017video, wang2018reconstruction, wang2018video, zhang2020object}, video title generation~\cite{amirian2021automatic,  zeng2016title, zhang2020comprehensive}, generic event
boundary captioning~\cite{wang2022geb+} and dense video captioning~\cite{krishna2017dense, wang2021end, zhou2018end}.
Most related to video chapter generation, the dense video captioning task requires temporally localizing and captioning all events in an untrimmed video.
In contrast, video chapter generation requires temporally \emph{segmenting} the video (i.e.~the start of the chapter $i+1$ is the end of chapter $i$, and the chapters cover the full video), and involves generating a chapter title that is substantially shorter than a video caption.
We study in more detail the transfer learning between these two tasks in Section~\ref{sec:dvc}.
Finally, the video chapter grounding task is related to temporal language grounding~\cite{hendricks17localizing, hendricks2018localizing, lei2020tvr, lei2021detecting, nan2021interventional, yang2022tubedetr, zhang2020span, zhang2020learning}.
However, we here focus on localizing a chapter starting point and not a start-end window.
Furthermore, most temporal language grounding methods represent the video only with visual inputs, while we also exhibit the benefits of using speech inputs for localizing chapters in videos (see Section~\ref{sec:vcgr}).

\section{\dataset{}: a large-scale dataset of user-chaptered videos}\label{sec:dataset}
Our goal is to build a large and diverse set of videos annotated with temporarily localized chapter information, consisting of chapter titles and chapter start times.
In detail, chapters are contiguous, non-overlapping segments, completely partitioning a video.
However manual annotation of chapters is time consuming and expensive and therefore hard to scale.
Hence we automatically scrape chapter information from videos available online, as explained in Section~\ref{sec:collection}.
Then, we perform several processing steps on this data, e.g., to extract speech transcripts, as described in Section~\ref{sec:processing}. 
The outcome is \dataset{}, a dataset of \dsetsize{} videos with \chaptersize{} chapter annotations provided by real users online.
Finally, we analyze \dataset{} in Section~\ref{sec:analysis}.
Details are given next. 

\subsection{Data collection}\label{sec:collection}
Since early 2020, YouTube users can create chapters for uploaded videos by annotating them in the YouTube description. 
The YouTube API, however, currently does not enable explicit search for user-chaptered videos.
Hence, our data collection procedure consists of:
(i) Collecting a large and diverse set of video candidates (characterized by their 11-character YouTube video ID), which do not necessarily contain user-annotated chapters;
(ii) For all video candidates, downloading the video description, automatically selecting videos with user-annotated chapters, extracting video chapters and downloading corresponding videos.
We next describe the individual steps in more detail.

\noindent \textbf{Video candidates.} 
We start from a large pool of video candidates built from the YT-Temporal-180M dataset~\cite{zellers2021merlot}, which was constructed to be more diverse than prior large video datasets such as HowTo100M~\cite{miech19howto100m}.
Note that while the released YT-Temporal-180M dataset consists of only 5M videos, the authors collected a larger set of candidates by using YouTube's recommendation algorithm to suggest related videos. We obtained this extended list of 92 million video IDs directly from the authors. 

\noindent \textbf{Extracting chapters from descriptions.} 
In the description, chapters typically constitute a block with consecutive lines following the format {\small \texttt{\color{greencode}``<Timestamp>: <Chapter Title>''}} or {\small \texttt{\color{greencode}``<Chapter Title>: <Timestamp>''}}, where the chapter title is written in free-form natural language and its corresponding start timestamp is written in \texttt{MM:SS} format.
The video should contain at least two timestamps listed in ascending order.
Hence we download the descriptions for all video candidates and use standard regular expression operations to verify whether a given description contains user-annotated chapters and extract them if so.
Note that some videos contain chapters that are automatically generated by YouTube algorithms, however, these generated chapters do not appear in the descriptions and, hence, are excluded by our procedure for data collection.
Also note that the video content is only downloaded for user-chaptered videos, which is convenient for both the downloading speed and storage constraints.
Finally, we obtain \dsetsize{} user-chaptered videos, making up 0.9\% of all video candidates.

\subsection{Data processing}\label{sec:processing}
We describe below how we process the previously obtained user-chaptered videos to facilitate building efficient video chapter generation models.
For reproducibility, we publicly release the resulting speech transcripts and the code for extracting visual features.

\noindent \textbf{ASR extraction.}
We observed that most user-chaptered videos contain speech.
Hence, for all videos, we extract speech transcripts aligned in time with the video content (ASR) by applying the Whisper-Large-V2 model~\cite{radford2022robust} on the audio track, using faster-whisper~\cite{fasterwhisper} backend for computational efficiency.
We found that the Whisper model provides higher-quality ASR compared to the YouTube API ASR service on several data samples from \dataset{}.
We further use WhisperX~\cite{bain2022whisperx} to derive accurate word-level timestamps which we use to segment the speech transcript into sentences.
For example, the Whisper-Large-V2 model extracts speech segments like \textit{“Right, we're gonna do the Synthetics Dirty Race. No we're not.
[...]
So we're gonna put two t-shirts and two pairs of jeans in the”} with timestamps 20.478s and 50.465s, and the corresponding first sentence output by WhisperX is \textit{“Right, we're gonna do the Synthetics Dirty Race.”} with timestamps 20.538s and 29.26s.

\noindent \textbf{Visual feature extraction.}
Training end-to-end deep learning models from RGB inputs on minutes-long videos is computationally expensive.
Hence we extract visual features with CLIP ViT-L/14 backbone~\cite{dosovitskiy2021an, radford2021learning} at resolution $224\times224$ pixels and 1 FPS.
This model has been trained to map images to text descriptions with a contrastive loss on 400M Web-scraped image-text pairs.

\begin{figure}[t]
\centering
\begin{subfigure}{.33\textwidth}
\includegraphics[width=\linewidth]{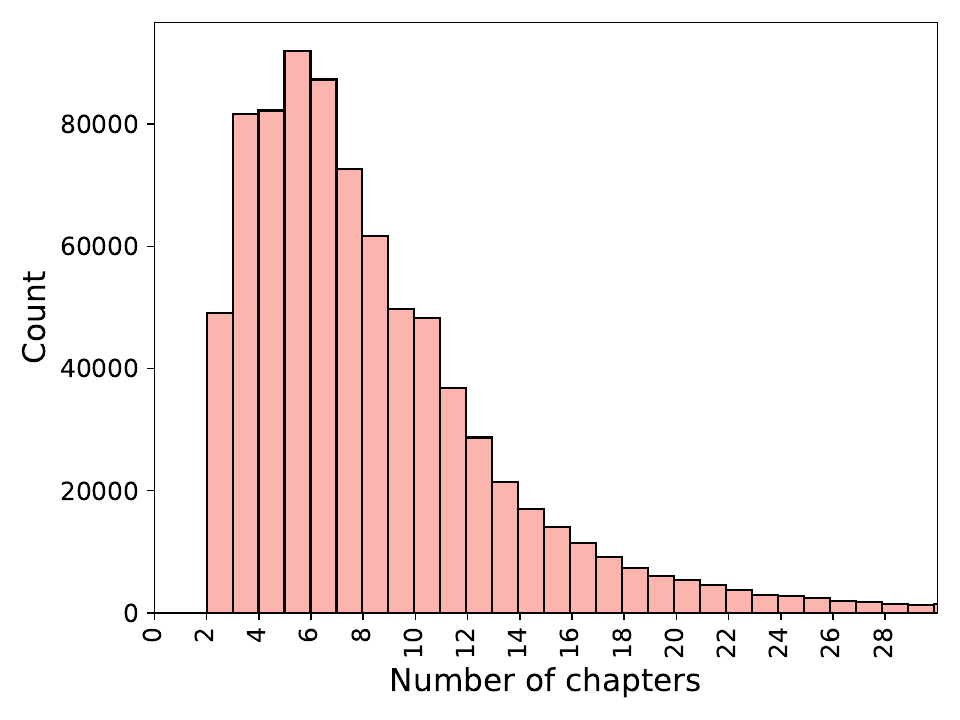}
\caption{Number of chapters per video}
\end{subfigure}%
\begin{subfigure}{.33\textwidth}
\includegraphics[width=\linewidth]{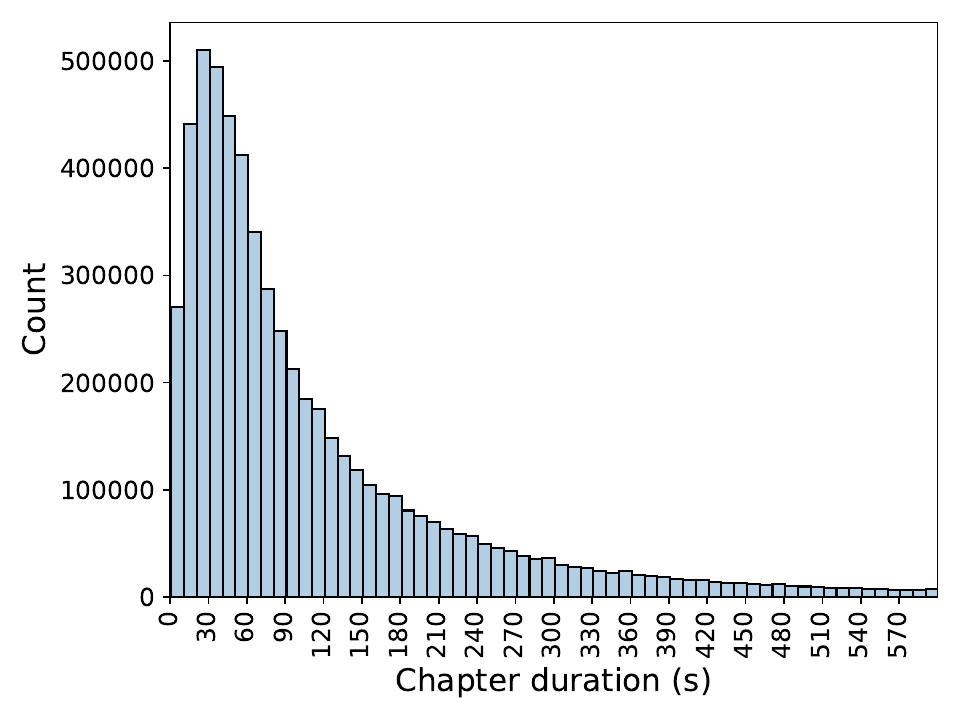}
\caption{Video chapter duration (s)}
\end{subfigure}%
\begin{subfigure}{.33\textwidth}
\includegraphics[width=\linewidth]{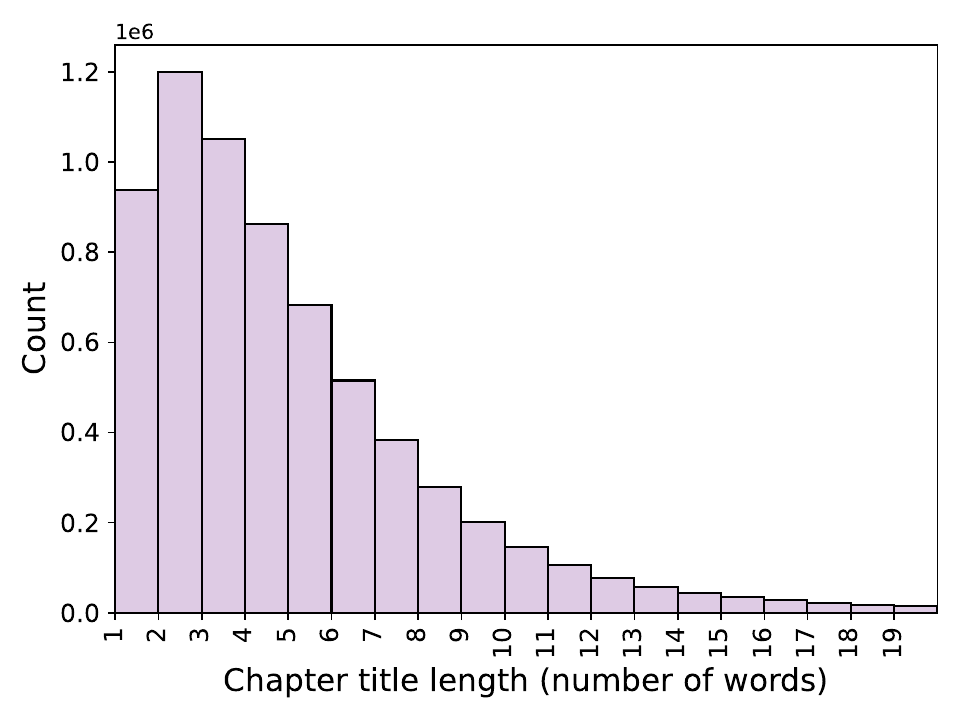}
\caption{Chapter title length (\# words)}
\end{subfigure}
\begin{subfigure}[valign=t]{.49\textwidth}
\includegraphics[width=\linewidth]{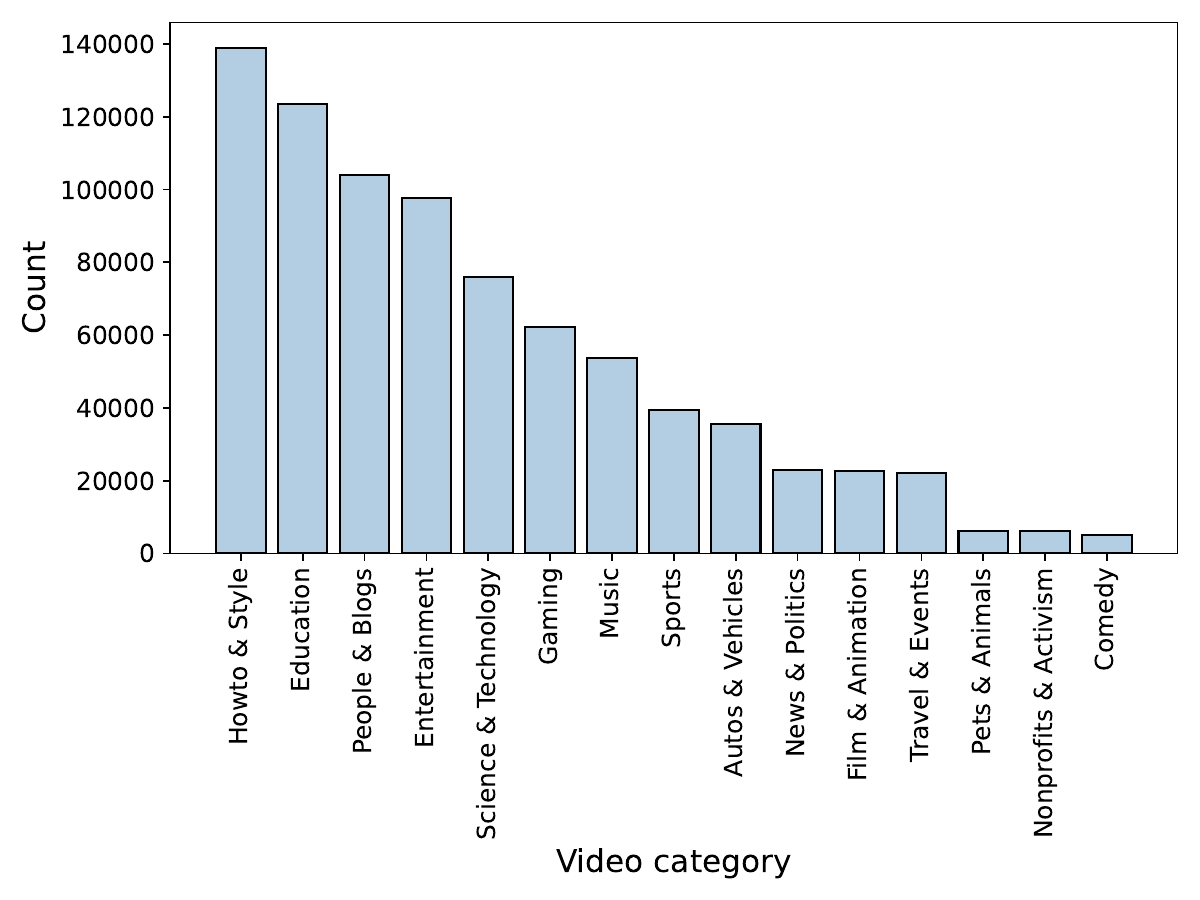}
\caption{Video category}
\end{subfigure}%
\begin{subfigure}[valign=t]{.49\textwidth}
\includegraphics[width=\linewidth]{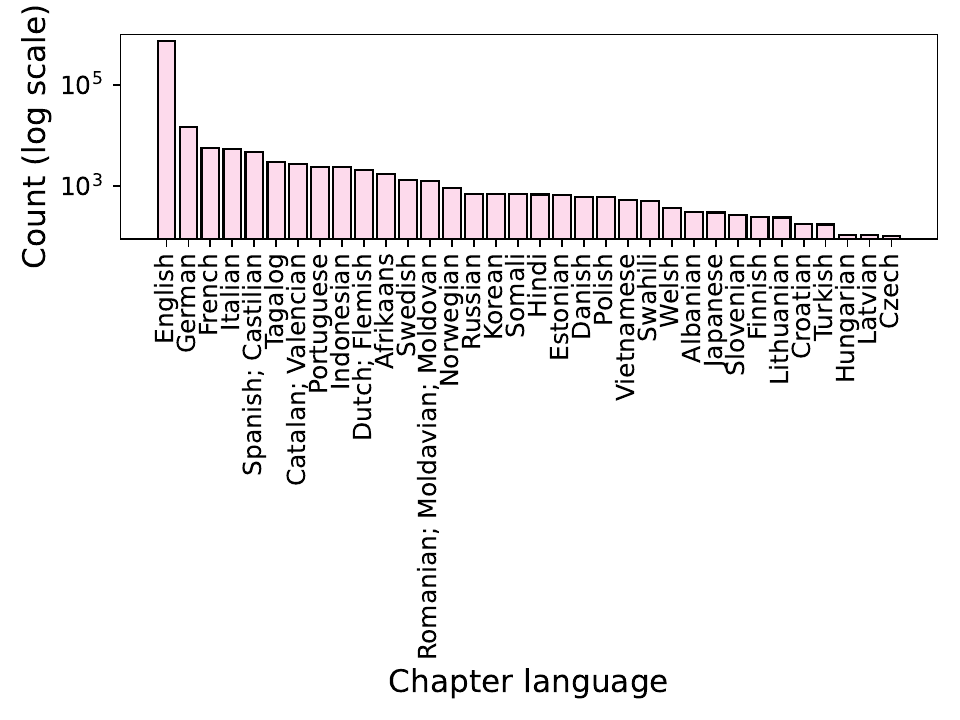}
\caption{Chapter languages}
\end{subfigure}
\caption{{\bf Statistics of the \dataset{} dataset.}}
\label{fig:stats}
\vspace{-.5cm}
\end{figure}

\subsection{Data analysis}\label{sec:analysis}
The result of the previously described pipeline is \dataset{},
a dataset of 817,076 user-chaptered videos containing 6,813,732 chapters in total.
We randomly split \dataset{} in training, validation, and testing splits with 801K, 8.2K, and 8.2K videos, respectively.
We analyze \dataset{} below and give examples of annotated videos, more statistics, as well as a datasheet in \if\sepappendix1{the Supplementary Material}\else{Appendix Sections~\ref{supp:addex},~\ref{supp:adddata}, and~\ref{supp:datasheet}, respectively}\fi.

\noindent \textbf{Statistics.}
\dataset{} is highly diverse and contains 4,894,855 distinct chapter titles.
On average, a video contains 8.3 chapters, start times of adjacent chapters are separated by 142.0s seconds, a chapter title contains 5.4 words and a video lasts 1354 seconds.
The most represented video category (in YouTube's glossary) is HowTo \& Style, making up 17.0\% of total videos.
The distributions for the number of chapters per video, the video chapter duration, the length of the chapter title, 
and the video category are illustrated in Figure~\ref{fig:stats}, and further show the diversity of \dataset{}, e.g., there are 12 different video categories with at least 20K videos in \dataset{}.

\noindent \textbf{ASR vs Chapters.}
97.3\% of videos in \dataset{} contain speech transcripts (ASR).
However, user-annotated chapters significantly differ from speech transcripts: on average, a video with ASR contains 269.8 speech sentences (vs 8.3 chapter titles), a speech sentence lasts 3.9 seconds (vs 142.0 seconds for chapters) in the video and contains 11.5 words (vs 5.4 words for chapters).

\noindent \textbf{Biases.}
Using the langdetect~\cite{langdetect} language detection tool, we find that 92.9\%/93.9\% of total videos in \dataset{} have their chapter titles/ASR in English.
However, as shown in Figure~\ref{fig:stats} (bottom right), the distribution of chapter languages includes a long tail of languages, e.g., 13 languages appear in more than 1K videos of \dataset{}.
We also use GenBit~\cite{sengupta2021genbit} to measure gender bias in the chapters and ASR.
We observe that the percentage of female/male/non-binary gendered words is 19.7\%/39.7\%/40.7\% for the chapters, and 11.6\%/35.6\%/52.8\% for the ASR.

\noindent \textbf{Ethical considerations.}
We employ several techniques to identify harmful visual or language content.
We use a classifier~\cite{schuhmannlaion} built on top of the previously extracted CLIP features to detect 
not-safe-for-work (NSFW) visual content (such as pornographic and sexualized content).
Moreover, we use a language model~\cite{Detoxify} to detect toxic content in chapter titles and speech transcripts.
These processes flag 5,716 (0.70\%) visually NSFW videos, 355 (0.04\%) videos with toxic chapter titles and 1,368 (0.17\%) videos with toxic ASR.
We assume the relatively low number of flagged videos is due to the regulations performed by the Web platform used to collect our dataset. 
Following~\cite{schuhmannlaion}, we refrain from removing these samples to encourage research in fields such as dataset curation and tag them instead.
Note that these automated filtering techniques are not perfect and that harmful content may pass.

\begin{table}[t]
\begin{center}
\setlength\tabcolsep{6pt}
\resizebox{.35\linewidth}{!}{	
\begin{tabular}{l|c}
Type of chapter titles & Percentage \\ 
\hline
Speech and visual & 49 \\
Audio and visual & 2 \\
Speech-only & 26 \\
Visual-only & 3 \\
Audio-only & 3 \\
\hline
Structure-only & 14 \\
\hline
Unrelated & 3 \\
\end{tabular}}
\end{center}
\caption{\textbf{Manual assessment of the informativeness of chapter titles in the \dataset{} dataset over a random sample of 100 videos.}
Video chapter titles can be based on speech and vision; audio and vision; vision, audio or speech alone; or only on the structure of the video (\textit{e.g.}~"step 1", "step 2" etc). 
In a small number of cases, video chapters are unrelated to the video content.
}
% \vspace{-.5cm}
\label{table:manual}
\end{table} 

\noindent \textbf{Manual assessment of the quality of annotations.}
While chapter titles are manually written and uploaded by real users, sometimes chapter titles are not informative about the content of the video at the corresponding timestamps. 
To assess the quality of chapter title annotations in our dataset, we inspected a random sample of 100 videos in \dataset{}. 
For each video, we checked if the titles are related to the content of the video chapter and if so which video modalities (ASR, visual or raw audio) they are related to, or if they only refer to the structure of the video (e.g. chapter titles like "step 1", "step 2" etc). Results are presented in Table~\ref{table:manual}, and show that 83\% of videos have chapters related to one or multiple modalities of the video, 14\% of videos have chapters only referring to the structure of the video, and 3\% of videos have chapters unrelated to the video content.

\section{Experiments}\label{sec:experiments}
In this Section, we present the results of models on \dataset{} for the full video chapter generation task in Section~\ref{sec:vcg}, the task of video chapter generation given ground-truth boundaries in Section~\ref{sec:vcggt} and the video chapter grounding task in Section~\ref{sec:vcgr}.
Finally, we study transfer learning from video chapter generation to dense video captioning tasks in Section~\ref{sec:dvc}.

\noindent \textbf{Evaluation metrics.}
To evaluate the quality of the generated chapter titles (without their positions), we use standard metrics used for visual captioning: BLEU~\cite{papineni2002bleu} (B), 
CIDEr~\cite{vedantam2015cider} (C), METEOR~\cite{banerjee2005meteor} (M) and ROUGE-L~\cite{lin2004rouge}~(RL).
To evaluate video chapter generation as a whole, including the locations of the generated chapters, we follow standard protocols used for dense video captioning, given the similar nature of the two tasks. We use the standard evaluation tool~\cite{krishna2017dense} which calculates matched pairs between generated events and the ground truth across IoU thresholds of \{0.3, 0.5, 0.7, 0.9\}, and compute captioning metrics over the matched pairs.
However, these metrics do not take into account the story of the video and give high scores to methods generating many redundant chapters.
Hence for an overall evaluation, we also use SODA\_c~\cite{fujita2020soda} (S) which first tries to find a temporally optimal matching between generated and reference chapters to capture the story of a video, then computes METEOR scores for the matching and derives F-measure scores from the METEOR scores to penalize redundant chapters. 
To separately evaluate chapter localization, we report the recall (R@Ks, R@K) and the precision (P@Ks, P@K) across various thresholds in terms of the distance to the ground-truth start time 
or IoU with the ground-truth start-end window.
We also report the average recall (R) and average precision (P) across IoU thresholds of \{0.3, 0.5, 0.7, 0.9\}.

\noindent \textbf{Implementation details.}
Unless stated otherwise, for all models, we use the speech transcripts (ASR) and visual features extracted as explained in Section~\ref{sec:processing}.
By default, each model is taken from the corresponding official implementation, and all model hyper-parameters are set according to the original papers.
We use the Adam optimizer~\cite{kingma15adam} for training and select the final model based on the best validation performance.
Our experiments are run on 8 NVIDIA A100 80GB GPUs.
More details are included in \if\sepappendix1{the Supplementary Material}\else{Appendix Section~\ref{supp:addimplem}}\fi.

\subsection{Video chapter generation}\label{sec:vcg}

In this Section, we study the task of video chapter generation that requires temporally segmenting the video and generating a chapter title for each segment.

\noindent \textbf{Models.}
For the video chapter segmentation subtask, we evaluate two zero-shot approaches (i.e., that are not trained on \dataset{}): speech text tiling~\cite{hearst1997text}, which detects subtopic shifts based on the analysis of lexical co-occurrence patterns, and a visual scene change detection algorithm~\cite{tomar2006converting} based on the sum of absolute differences.
To derive zero-shot baselines for the full video chapter generation task, we combine text tiling and shot detection with various alternatives that can generate text given text or visual input: a random baseline that predicts a random speech sentence spoken inside the predicted boundaries, LLaMA-7B~\cite{touvron2023llama} (prompted to summarize the speech transcript spoken inside the predicted boundaries) and BLIP-2~\cite{li2023blip} (prompted to describe the middle video frame of the predicted segment).
Finally, we also train and evaluate two state-of-the-art end-to-end dense video captioning models on \dataset{}: PDVC~\cite{wang2021end} which consists of a visual-only DETR-style~\cite{carion2020end} architecture and Vid2Seq~\cite{yang2023vid2seq} which is a multi-modal sequence-to-sequence model pretrained on the C4 text corpus~\cite{raffel2020exploring} and on narrated videos with ASR (\textit{e.g.}, YT-Temporal-1B~\cite{zellers2022merlot}).
For Vid2Seq, we also report zero-shot results after pretraining on narrated videos without finetuning on \dataset{}.

\begin{table}[t]
\begin{center}
\setlength\tabcolsep{3pt}
\resizebox{1.\linewidth}{!}{
\begin{tabular}{l|ccc|cccccccc}
Method & Modalities & Pretraining Data & Finetuned & S & B1 & B2 & B3 & B4 & C & M & RL \\
\hline
Text tiling~\cite{hearst1997text} + Random & Speech & $\emptyset$ & \xmark
& 0.4 & 0.6 & 0.2 & 0.1 & 0.0 & 0.8 & 0.7 & 0.6 \\
Text tiling~\cite{hearst1997text} + LLaMA~\cite{touvron2023llama} & Speech & Text mixture & \xmark
& 0.2 & 0.4 & 0.1 & 0.1 & 0.0 & 0.5 & 0.3 & 0.4 \\
Shot detect~\cite{tomar2006converting} + BLIP-2~\cite{li2023blip} & Visual & 129M image-texts & \xmark
& 0.6 & 0.7 & 0.3 & 0.1 & 0.1 & 0.2 & 0.6 & 0.8 \\
Vid2Seq~\cite{yang2023vid2seq} & Speech+Visual & C4 + HowTo100M & \xmark
& 0.1 & 0.1 & 0.0 & 0.0 & 0.0 & 0.1 & 0.1 & 0.1 \\
\hline
PDVC~\cite{wang2021end} & Visual & $\emptyset$ & \cmark
& 6.8 & 9.4 & 3.7 & 1.4 & 0.9 & 35.8 & 9.4 & 11.4 \\
Vid2Seq~\cite{yang2023vid2seq} & Speech & C4 & \cmark
& 10.2 & 9.5 & 6.7 & 4.0 & 2.7 & 48.8 & 8.5 & 11.0 \\
Vid2Seq~\cite{yang2023vid2seq} & Speech & C4 + HowTo100M & \cmark
& 10.5 & 9.9 & 7.0 & 4.2 & 2.9 & 50.7 & 8.7 & 11.4 \\
Vid2Seq~\cite{yang2023vid2seq} & Visual & C4 & \cmark
& 3.1 & 2.3 & 1.5 & 0.6 & 0.5 & 10.9 & 2.2 & 2.9 \\
Vid2Seq~\cite{yang2023vid2seq} & Visual & C4 + HowTo100M & \cmark
& 5.5 & 4.5 & 2.8 & 1.2 & 0.9 & 21.1 & 4.1 & 5.5 \\
Vid2Seq~\cite{yang2023vid2seq} & Speech+Visual & C4 & \cmark
& 10.6 & 9.9 & 7.0 & 4.2 & 2.8 & 51.3 & 8.8 & 11.6 \\
Vid2Seq~\cite{yang2023vid2seq} & Speech+Visual & C4 + HowTo100M & \cmark
& \textbf{11.4} & \textbf{10.9} & \textbf{7.7} & \textbf{4.6} & \textbf{3.1} & \textbf{55.7} & \textbf{9.5} & \textbf{12.6} \\
\end{tabular}
}
\vspace{0.1cm}
\caption{\textbf{Video chapter generation (global metrics) on \dataset{} test set.} 
Here, finetuned refers to finetuning on the \dataset{} train set, and speech refers to transcribed speech (ASR).}
\label{table:vcg}
% \vspace{-0.5cm}
\end{center}
\end{table}

\begin{table}[t]
\begin{center}
\setlength\tabcolsep{2pt}
\resizebox{1.\linewidth}{!}{
\begin{tabular}{l|ccc|cccccccc}
Method & Modalities & Pretraining Data & Finetuned & R@5s & R@3s & R@0.5 & R@0.7 & P@5s & P@3s & P@0.5 & P@0.7 \\
\hline
Text tiling~\cite{hearst1997text} & Speech & $\emptyset$ & \xmark
& 9.4 & 5.8 & 23.6 & 8.9 & 12.6 & 7.9 & 26.0 & 8.8 \\
Shot detect~\cite{tomar2006converting} & Visual & $\emptyset$ & \xmark
& 31.2 & 27.4 & 24.9 & 12.5 & 33.2 & 29.7 & 18.0 & 8.7 \\
Vid2Seq~\cite{yang2023vid2seq} & Speech+Visual & C4 + HowTo100M & \xmark
& 10.7 & 9.5 & 5.8 & 0.2 & 23.3 & 18.5 & 1.9 & 0.8 \\
\hline
PDVC~\cite{wang2021end} & Visual & $\emptyset$ & \cmark
& 21.1 & 17.8 & 31.2 & 22.5 & \textbf{45.3} & \textbf{40.2} & \textbf{47.2} & \textbf{26.9} \\
Vid2Seq~\cite{yang2023vid2seq} & Speech & C4 & \cmark
& \textbf{37.8} & \textbf{29.5} & 44.6 & 26.1 & 29.0 & 23.0 & 38.0 & 23.4 \\
Vid2Seq~\cite{yang2023vid2seq} & Speech & C4 + HowTo100M & \cmark
& 36.7 & 28.9 & 46.5 & 27.2 & 29.5 & 23.3 & 40.4 & 24.8 \\
Vid2Seq~\cite{yang2023vid2seq} & Visual & C4 & \cmark
& 35.3 & 26.4 & 23.6 & 8.7 & 17.9 & 13.6 & 17.2 & 7.1 \\
Vid2Seq~\cite{yang2023vid2seq} & Visual & C4 + HowTo100M & \cmark
& 33.5 & 25.0 & 33.0 & 14.5 & 19.5 & 14.7 & 26.2 & 12.5 \\
Vid2Seq~\cite{yang2023vid2seq} & Speech+Visual & C4 & \cmark
& 36.3 & 28.6 & 45.8 & 26.9 & 29.9 & 23.8 & 40.9 & 24.9 \\
Vid2Seq~\cite{yang2023vid2seq} & Speech+Visual & C4 + HowTo100M & \cmark
& 36.4 & 28.5 & \textbf{48.2} & \textbf{28.5} & 30.3 & 24.0 & 43.1 & 26.4 \\
\end{tabular}
}
\vspace{0.1cm}
\caption{\textbf{Video chapter generation (segmentation metrics) on \dataset{} test set}.}
\label{table:vcs}
% \vspace{-0.5cm}
\end{center}
\end{table}

\noindent \textbf{Implementation details.}
We use the text tiling implementation from the NLTK library~\cite{bird2009natural} which tokenizes the text into pseudosentences of size 50.
We use the shot detection software from the FFMPEG library~\cite{tomar2006converting} with a confidence threshold of 0.7.
For BLIP-2, we use the 3.4B-parameter variant with FLAN-T5-XL~\cite{wei2021finetuned} and CLIP ViT-L/14~\cite{radford2021learning, dosovitskiy2021an}.
We reimplement Vid2Seq~\cite{yang2023vid2seq} (originally released in Jax) in PyTorch, use T5-Base pretrained on C4~\cite{raffel2020exploring} for initialization and pretrain Vid2Seq on HowTo100M~\cite{miech19howto100m}.
More details are included in \if\sepappendix1{the Supplementary Material}\else{Appendix Section~\ref{supp:addimplem}}\fi.

\noindent \textbf{Results.}
We report the results for video chapter generation using global metrics and localization-only metrics in Tables~\ref{table:vcg} and~\ref{table:vcs}, respectively.
We observe that models trained on \dataset{} outperform zero-shot baselines, 
demonstrating the effectiveness of training on \dataset{}.
In particular, PDVC~\cite{wang2021end} has the best precision and Vid2Seq~\cite{yang2023vid2seq} achieves the best results in terms of overall generation and recall.
We also find that Vid2Seq's speech-only mode outperforms its visual-only mode and that using both speech and visual inputs leads to the best performance.
This demonstrates that video chapter generation is a multi-modal task.
Finally, we observe that pretraining using ASR in narrated videos from HowTo100M~\cite{miech19howto100m} improves the video chapter generation performance of the Vid2Seq model.
Specifically, pretraining on HowTo100M is more beneficial for vision-aware models than for the speech-only model.

\noindent \textbf{Qualitative examples.} See \if\sepappendix1{the Supplementary Material}\else{Appendix Section~\ref{supp:addpred}}\fi.

\begin{table}[t]
\begin{center}
\setlength\tabcolsep{3pt}
\resizebox{.8\linewidth}{!}{
\begin{tabular}{l|ccc|ccccccc}
Method & Modalities & Pretraining Data & Finetuned & B1 & B2 & B3 & B4 & C & M & RL \\
\hline
Random & Speech & $\emptyset$ & \xmark
& 2.4 & 1.3 & 0.9 & 0.7 & 10.4 & 2.2 & 4.4 \\
LLaMA~\cite{touvron2023llama} & Speech & Text mixture & \xmark
& 0.0 & 0.0 & 0.0 & 0.0 & 0.0 & 0.1 & 0.2 \\
BLIP-2~\cite{li2023blip} & Visual & 129M image-texts & \xmark
& 3.1 & 1.5 & 0.9 & 0.7 & 12.4 & 2.2 & 4.5 \\
Vid2Seq~\cite{yang2023vid2seq} & Speech+Visual & C4 + HowTo100M & \xmark
& 2.0 & 1.2 & 0.9 & 0.6 & 0.9 & 0.3 & 0.6 \\
\hline
Vid2Seq~\cite{yang2023vid2seq} & Speech & C4 + HowTo100M & \cmark
& 21.0 & 15.5 & 12.1 & 10.0 & 105.3 & 11.5 & 24.5 \\
Vid2Seq~\cite{yang2023vid2seq} & Visual & C4 + HowTo100M & \cmark
& 10.1 & 5.6 & 3.5 & 2.4 & 47.1 & 5.1 & 14.7 \\
Vid2Seq~\cite{yang2023vid2seq} & Speech+Visual & C4 & \cmark
& 21.6 & 15.7 & 12.3 & 10.0 & 110.8 & 11.5 & 26.0 \\
Vid2Seq~\cite{yang2023vid2seq} & Speech+Visual & C4 + HowTo100M & \cmark
& \textbf{23.5} & \textbf{17.2} & \textbf{13.4} & \textbf{11.0} & \textbf{120.5} & \textbf{12.6} & \textbf{28.3} \\
\end{tabular}
}
\vspace{0.1cm}
\caption{\textbf{Chapter title generation given ground-truth boundaries on \dataset{} test set.}}
\label{table:vcggt}
% \vspace{-0.5cm}
\end{center}
\end{table}

\subsection{Video chapter generation given ground-truth boundaries}\label{sec:vcggt}

In this Section, we study the task of generating chapter titles provided correct temporal boundaries of video chapters. This task is a simplification of the previously studied task where we assume perfect temporal segmentation.
We adopt the same models and implementation details as previously introduced in Section~\ref{sec:vcg}.

\noindent \textbf{Results.}
We report results for video chapter generation given ground-truth boundaries in Table~\ref{table:vcggt}.
Similar to the full video chapter generation task, we observe that solving the task without training on \dataset{} is hard.
Indeed, LLaMA~\cite{touvron2023llama} struggles to summarize the speech content into a chapter title and underperforms the random baseline.
Furthermore, BLIP-2~\cite{li2023blip} slightly improves over the random baseline.
In addition, Vid2Seq~\cite{yang2023vid2seq} in zero-shot mode underperforms the random baseline due to the large domain gap between ASR and chapter titles (see Section~\ref{sec:analysis}).
In comparison, the performance of models trained on \dataset{} is significantly higher.
Moreover, Vid2Seq's speech-only mode outperforms its visual-only mode, and using both speech and visual inputs is beneficial, confirming the benefit of multi-modal reasoning for the task of generating chapter titles.
Finally, pretraining on narrated videos from HowTo100M~\cite{miech19howto100m} improves the performance of the Vid2Seq model on \dataset{}.

\begin{table}[t]
\begin{center}
\setlength\tabcolsep{1pt}
\resizebox{1.\linewidth}{!}{
\begin{tabular}{l|ccc|cccccccc}
Method & Modalities & Pretraining Data & Finetuned & R@10s & R@5s & R@3s & R@1s & R@0.3 & R@0.5 & R@0.7 & R@0.9 \\
\hline
Random & Speech & $\emptyset$ & \xmark
& 3.1 & 1.8 & 1.2 & 0.6 & 0.7 & 0.3 & 0.1 & 0.0 \\
BERT~\cite{bert18} & Speech & BookCorpus + Wikipedia & \xmark
& 9.0 & 6.8 & 5.4 & 2.9 & 0.6 & 0.3 & 0.1 & 0.0 \\
CLIP~\cite{radford2021learning} & Visual & 400M image-texts & \xmark
& 8.1 & 5.2 & 3.7 & 1.4 & 10.7 & 5.2 & 2.3 & 0.5 \\
Moment-DETR~\cite{lei2021detecting} & Visual & 5.4K narrated videos~\cite{lei2021detecting} & \xmark
& 3.2 & 1.6 & 1.1 & 0.5 & 11.3 & 3.6 & 0.8 & 0.1 \\
\hline
Moment-DETR~\cite{lei2021detecting} & Visual & $\emptyset$ & \cmark
& \textbf{21.8} & \textbf{15.5} & \textbf{12.4} & \textbf{8.3} & \textbf{37.4} & \textbf{27.3} & \textbf{17.6} & \textbf{6.4} \\
\end{tabular}
}
\vspace{0.1cm}
\caption{\textbf{Video chapter grounding on \dataset{} test set.}}
\label{table:vcgr}
% \vspace{-0.6cm}
\end{center}
\end{table}

\subsection{Video chapter grounding}\label{sec:vcgr}

In this Section, we study the task of video chapter grounding that requires a model to temporally localize a chapter start time (or start-end window) given an annotated chapter title (query).
Hence, compared to the video chapter generation task, we here assume chapter titles to be given and focus on the temporal chapter localization only.

\noindent \textbf{Models.}
We evaluate three zero-shot alternatives: a random baseline that randomly picks the timestamps of a speech sentence in the video, a BERT~\cite{bert18} baseline that picks the timestamps of the speech sentence that has the closest text embedding with the queried chapter title, and a CLIP~\cite{radford2021learning} baseline picking the frames where the query-frame similarity score drops from the highest scoring frame by a certain threshold $\epsilon$.
We also train and evaluate on \dataset{} a state-of-the-art end-to-end video grounding model: Moment-DETR~\cite{lei2021detecting} which is designed for moment retrieval based on visual inputs.
Furthermore, we report zero-shot performance of Moment-DETR obtained with the model checkpoint from~\citet{lei2021detecting} pretrained on 5.4K narrated videos with ASR from the QVHighlights dataset~\cite{lei2021detecting}.

\noindent \textbf{Implementation details.} 
We use the \texttt{[CLS]} token sequence embedding for the BERT baseline and a threshold of $\epsilon=0.05$ for the CLIP baseline.
More details are provided in \if\sepappendix1{the Supplementary Material}\else{Appendix Section~\ref{supp:addimplem}}\fi.

\noindent \textbf{Results.}
We report results for the video chapter grounding task in Table~\ref{table:vcgr}.
We first observe that the simple zero-shot baselines based on ASR can decently find start times, but struggle to predict start-end windows due to the important domain gap between ASR and video chapters (see Section~\ref{sec:analysis}).
The CLIP~\cite{radford2021learning} baseline slightly underperforms the BERT baseline~\cite{bert18} at retrieving start times, but is much better at finding start-end windows. 
Furthermore, the Moment-DETR model~\cite{lei2021detecting} trained on \dataset{} outperform the zero-shot baselines for both localization of start times and start-end windows, which further demonstrates the effectiveness of training on \dataset{}.
Finally, we note that Moment-DETR cannot handle speech inputs, but hope that our results showing the benefit of this modality on other tasks in \dataset{} will foster research in the localization of language queries in untrimmed videos using multi-modal inputs (vision and speech transcripts).

\begin{table}[t]
\begin{center}
\setlength\tabcolsep{3pt}
\resizebox{1.\linewidth}{!}{
\begin{tabular}{l|cc|ccccc|ccccc}
\multirow{2}{*}{Method} & \multirow{2}{*}{Modalities} & \multirow{2}{*}{Pretraining Data}
& \multicolumn{5}{c|}{YouCook2 (val)}
& \multicolumn{5}{c}{ViTT (test)} \\
& & & S & C & M & R & P
& S & C & M & R & P \\
\hline
PDVC~\cite{wang2021end} & V & $\emptyset$
& 4.4 & 22.7 & 4.7 & --- & ---
& --- & --- & --- & --- & --- \\
E2ESG~\cite{zhu2022end} & T+V & C4 + WikiHow
& --- & 25.0 & 3.5 & 20.7 & 20.6
& --- & 25.0 & 8.1 & 32.2 & 32.1 \\
Vid2Seq~\cite{yang2023vid2seq} & T+V
& C4 + HTM & 8.3 & 48.3 & 9.5 & 27.1 & 27.0
& --- & --- & --- & --- & --- \\
Vid2Seq~\cite{yang2023vid2seq} & T+V
& C4 + YT-Temporal-1B & 7.9 & 47.1 & 9.3 & 27.9 & 27.8
& 13.5 & 43.5 & 8.5 & 42.6 & 46.2 \\
\hline
PDVC$^\dag$ & V & $\emptyset$
& 4.8 & 28.8 & 5.8 & 22.6 & 33.1
& 9.4 & 40.6 & \textbf{16.5} & 19.2 & 37.4 \\
PDVC$^\dag$ & V & VC (Chap.)
& 5.9 & 34.7 & 7.5 & 28.8 & \textbf{36.4}
& 10.1 & 41.5 & 16.1 & 21.3 & 37.2 \\
Vid2Seq$^\dag$ & T+V & C4 + HTM
& 8.6 & 53.2 & 10.5 & 29.2 & 26.2
& 14.1 & 44.8 & 8.7 & 43.8 & 44.5 \\
Vid2Seq$^\dag$ & T+V & C4 + VC (ASR+Chap.)
& 9.8 & 62.9 & 11.7 & 32.5 & 30.1
& \textbf{15.1} & \textbf{50.9} & 9.6 & 45.1 & 46.7 \\
Vid2Seq$^\dag$ & T+V & C4 + HTM + VC (ASR)
& 8.4 & 50.1 & 10.3 & 29.7 & 26.3
& 14.3 & 45.6 & 8.8 & 43.7 & 44.9 \\
Vid2Seq$^\dag$ & T+V & C4 + HTM + 1\% of VC (ASR+Chap)
& 8.8 & 52.7 & 10.4 & 29.3 & 27.6
& 13.5 & 41.6 & 8.2 & 44.7 & 42.1 \\
Vid2Seq$^\dag$ & T+V & C4 + HTM + 10\% of VC (ASR+Chap.)
& 9.9 & 63.9 & 12.1 & 32.4 & 31.4
& 14.5 & 47.4 & 9.2 & 45.3 & 45.9 \\
Vid2Seq$^\dag$ & T+V & C4 + HTM + VC (ASR+Chap.)
& \textbf{10.3} & \textbf{67.2} & \textbf{12.3} & \textbf{34.0} & 31.2
& 15.0 & 50.0 & 9.5 & \textbf{45.5} & \textbf{46.9} \\
\end{tabular}
}
\vspace{0.1cm}
\caption{\textbf{Comparison with the state of the art on the YouCook2 and ViTT dense video captioning benchmarks.}
T: Transcribed speech, V: Visual, HTM: HowTo100M~\cite{miech19howto100m}, VC: \dataset{}, Chap.: Chapters.
$^\dag$ denote results of our experiments.}
\label{table:dvc}
% \vspace{-0.8cm}
\end{center}
\end{table}

\begin{table}[t]
\begin{center}
\setlength\tabcolsep{2pt}
\resizebox{1.\linewidth}{!}{
\begin{tabular}{l|cc|ccccc|ccccc}
\multirow{2}{*}{Method} & \multirow{2}{*}{Modalities} & \multirow{2}{*}{Pretraining Data}
& \multicolumn{5}{c|}{YouCook2 (val)}
& \multicolumn{5}{c}{ViTT (test)} \\
& & & S & C & M & R & P
& S & C & M & R & P \\
\hline
Text tiling~\cite{hearst1997text} + Random & T & $\emptyset$
& 0.3 & 0.9 & 0.3 & 3.8 & 6.6
& 0.3 & 0.6 & 0.6 & 11.6 & 24.4 \\
Text tiling~\cite{hearst1997text} + LLaMA~\cite{touvron2023llama} & T & Text mixture
& 0.2 & 0.6 & 0.2 & 3.8 & 6.6
& 0.2 & 0.6 & 0.5 & 11.6 & 24.4 \\
Shot detect~\cite{tomar2006converting} + BLIP-2~\cite{li2023blip} & V & 129M image-texts
& 0.6 & 1.0 & 0.5 & 8.9 & 5.5
& 0.2 & 0.1 & 0.2 & 3.1 & 13.7 \\
\hline
Vid2Seq~\cite{yang2023vid2seq} & V & C4 + VC (ASR)
& 0.0 & 0.0 & 0.0 & 0.0 & 0.0
& 0.0 & 0.0 & 0.0 & 0.2 & 0.8 \\
Vid2Seq~\cite{yang2023vid2seq} & V & C4 + VC (Chap.)
& 0.7 & 1.1 & 0.5 & 21.3 & 8.6
& 1.5 & 1.9 & 0.6 & 18.9 & 10.4 \\
Vid2Seq~\cite{yang2023vid2seq} & T+V & C4 + HTM
& 0.0 & 0.1 & 0.0 & 0.5 & 0.6
& 0.0 & 0.0 & 0.0 & 0.5 & 1.0 \\
Vid2Seq~\cite{yang2023vid2seq} & T+V & C4 + VC (ASR)
& 0.1 & 0.1 & 0.0 & 1.1 & 0.9
& 0.0 & 0.0 & 0.0 & 0.7 & 0.6 \\
Vid2Seq~\cite{yang2023vid2seq} & T+V & C4 + VC (Chap.)
& 0.1 & 0.2 & 0.1 & 0.7 & 1.4
& 0.7 & 1.1 & 0.3 & 14.3 & 12.8 \\
Vid2Seq~\cite{yang2023vid2seq} & T+V & C4 + VC (ASR+Chap.)
& 3.2 & 10.2 & 2.9 & 20.6 & 19.7 
& \textbf{9.1} & \textbf{30.2} & \textbf{6.7} & \textbf{33.8} & \textbf{40.8} \\
Vid2Seq~\cite{yang2023vid2seq} & T+V & C4 + HTM + VC (ASR)
& 0.0 & 0.1 & 0.0 & 1.2 & 0.9
& 0.0 & 0.0 & 0.0 & 0.8 & 0.7 \\
Vid2Seq~\cite{yang2023vid2seq} & T+V & C4 + HTM + 1\% of VC (ASR+Chap.)
& 2.7 & 7.2 & 2.1 & 18.1 & 17.3
& 5.5 & 15.5 & 4.3 & 31.3 & 37.1 \\
Vid2Seq~\cite{yang2023vid2seq} & T+V & C4 + HTM + 10\% of VC (ASR+Chap.)
& 3.2 & 11.5 & 3.0 & 19.4 & 19.2
& 6.4 & 21.6 & 5.3 & 31.0 & 38.2 \\
Vid2Seq~\cite{yang2023vid2seq} & T+V & C4 + HTM + VC (ASR+Chap.)
& \textbf{3.9} & \textbf{13.3} & \textbf{3.4} & \textbf{22.3} & \textbf{20.1} 
& 9.0 & 28.0 & 6.5 & 33.7 & 40.1 \\
\end{tabular}
}
\vspace{0.1cm}
\caption{\textbf{Zero-shot dense video captioning on the YouCook2 and ViTT benchmarks.}
T: Transcribed speech, V: Visual, HTM: HowTo100M~\cite{miech19howto100m}, VC: \dataset{}, Chap.: Chapters.}
\label{table:zsdvc}
% \vspace{-0.8cm}
\end{center}
\end{table}

\subsection{Transfer learning on dense video captioning}\label{sec:dvc}
In this Section, we investigate the pretraining of video-language models on our new \dataset{}. 
To this end, we adopt video chapter generation models trained on \dataset{} (see Section~\ref{sec:vcg}) to the tasks of dense video captioning with or without finetuning.

\noindent \textbf{Datasets.}
We use two dense video captioning datasets.
\textbf{YouCook2}~\cite{zhou18towards} has 2K untrimmed videos of cooking procedures.
On average, each video lasts 320s and is annotated with 7.7 temporally-localized sentences.
\textbf{ViTT}~\cite{huang2020multimodal} was created to better reflect the distribution of instructional videos in the wild compared to YouCook2, and consists of 8K untrimmed instructional videos.
On average, each video lasts 250s and is annotated with 7.1 temporally-localized short tags.
For both datasets, we extract speech transcripts and visual features as described in Section~\ref{sec:processing}, and follow the standard splits for training, validation and testing.
Note that we only use videos available on YouTube at the time of the work, resulting in 10 to 20\% less videos than in the original datasets.

\noindent \textbf{Implementation details.}
See Section~\ref{sec:vcg} and \if\sepappendix1{the Supplementary Material}\else{Appendix Section~\ref{supp:addimplem}}\fi.

\noindent \textbf{Results after finetuning.}
In Table~\ref{table:dvc}, we show that pretraining for video chapter generation on \dataset{} greatly improves the downstream dense video captioning performance compared to training from scratch or pretraining only with ASR data as done in previous work~\cite{yang2023vid2seq}.
We also find that pretraining both on HowTo100M~\cite{miech19howto100m} and \dataset{} results in the best overall performance.
In particular, the Vid2Seq model pretrained on both HowTo100M and \dataset{} largely improves the state of the art on both the YouCook2 and ViTT benchmarks.
In detail, on the YouCook2 benchmark, in the setting with C4 + HowTo100M pretraining, we observe that a boost of about 4.9 points in CIDEr is obtained with our reimplementation of Vid2Seq, and that 14.0 additional points in CIDEr are obtained by pretraining on \dataset{}.
Finally, we report the results of the Vid2Seq model after pretraining on different fractions of \dataset{} for a fixed number of iterations.
We construct these subsets such that larger subsets include the smaller ones.
These results suggest that the scale of the chapter dataset is an important factor in the downstream dense video captioning performance.
We conclude that \dataset{} opens a promising avenue for multi-modal pretraining.
We further show qualitative examples of dense video captioning in \if\sepappendix1{the Supplementary Material}\else{Appendix Section~\ref{supp:addpred}}\fi.

\noindent \textbf{Zero-shot dense video captioning.}
In Table~\ref{table:zsdvc}, we report results obtained by directly applying video chapter generation models trained on \dataset{} for dense video captioning without finetuning for this task.
As far as we know, our work is the first to explore this challenging zero-shot setting where no manual annotation of dense video captions is used for training.
The Vid2Seq model trained only using ASR data underperforms the random baseline, due to the large domain difference between speech transcripts and dense captions~\cite{yang2023vid2seq}.
In the visual-only setting, the variant trained on chapter annotations is better than the variant trained on ASR annotations.
In the visual+speech settings, only using chapter annotations does not perform well, as training only on chapters (i.e., without speech) does not enable the model to learn how to use the input speech modality at inference.
However, using both ASR and chapter annotations results in a largely better zero-shot dense video captioning performance and outperforms all baselines not trained on \dataset{}, demonstrating the complementary nature of the ASR and chapters annotations.
Finally, we also observe the benefits of increasing the size of the pretraining dataset of chapters in this setting.

\section{Conclusion, Limitations, and Societal Impacts}\label{sec:conclusion}
In this work, we presented \dataset{}, a large-scale dataset of user-chaptered videos.
Furthermore, we evaluated a variety of baselines on the tasks of video chapter generation with and without ground-truth boundaries and video chapter grounding.
Finally, we investigated the potential of \dataset{} for pretraining video-language models and demonstrated improved performance on the dense video captioning tasks.
\dataset{} thus provides a new resource to the research community that can be used both as a benchmark for the video chapter generation tasks and as a powerful means for pretraining generic video-language models.

\noindent \textbf{Limitations.}
As it is derived from YT-Temporal-180M~\cite{zellers2021merlot}, \dataset{} inherits the biases in the distribution of video categories reflected in this dataset.

\noindent \textbf{Societal Impacts.}
The development of video chapter generation models might facilitate potentially harmful downstream applications, e.g., video surveillance.
Moreover, models trained on \dataset{} might reflect biases present in videos from YouTube. 
It is important to keep this in mind when deploying, analysing and building upon these models.

\section*{Acknowledgements}

This work was granted access to the HPC resources of IDRIS under the allocation 2023-A0131011670 made by GENCI. 
The work was funded by Antoine Yang's Google PhD fellowship, the French government under management of Agence Nationale de la Recherche as part of the "Investissements d'avenir" program, reference ANR-19-P3IA-0001 (PRAIRIE 3IA Institute), the Louis Vuitton ENS Chair on Artificial Intelligence, the European Regional Development Fund under project IMPACT (reg.\ no.\ CZ.02.1.01/0.0/0.0/15 003/0000468).
We thank Jack Hessel and Rémi Lacroix for helping with collecting the dataset, and Antoine Miech for interesting discussions.

\bibliographystyle{plainnat}  
\bibliography{egbib}

\clearpage \newpage
\appendix

\section*{Appendix}
In this \if\sepappendix1{Supplementary Material}\else{Appendix}\fi, we present the following items:
\begin{itemize}
\item[\textit{(i)}] Additional examples from our \dataset{} dataset (Section~\ref{supp:addex}).
\item[\textit{(ii)}] Qualitative examples of video chapter generation and dense video caption prediction (Section~\ref{supp:addpred}).
\item[\textit{(iii)}] Additional data analysis of our \dataset{} dataset (Section~\ref{supp:adddata}).
\item[\textit{(iv)}] Additional implementation details (Section~\ref{supp:addimplem}).
\item[\textit{(v)}] Video chapter generation results split by language (Section~\ref{supp:addres}).
\item[\textit{(vi)}] A datasheet~\cite{gebru2021datasheets} for \dataset{} (Section~\ref{supp:datasheet}).
Note that in this datasheet, the hosting, licensing, and maintenance plan of \dataset{} is presented.
\end{itemize}
Note that our code, models and the \dataset{} dataset can be found on our website~\cite{chapterswebpage}.

\appendix

\section{Additional examples from \dataset{}}\label{supp:addex}
\begin{figure*}[tp]
\centering
\includegraphics[width=1.\linewidth]{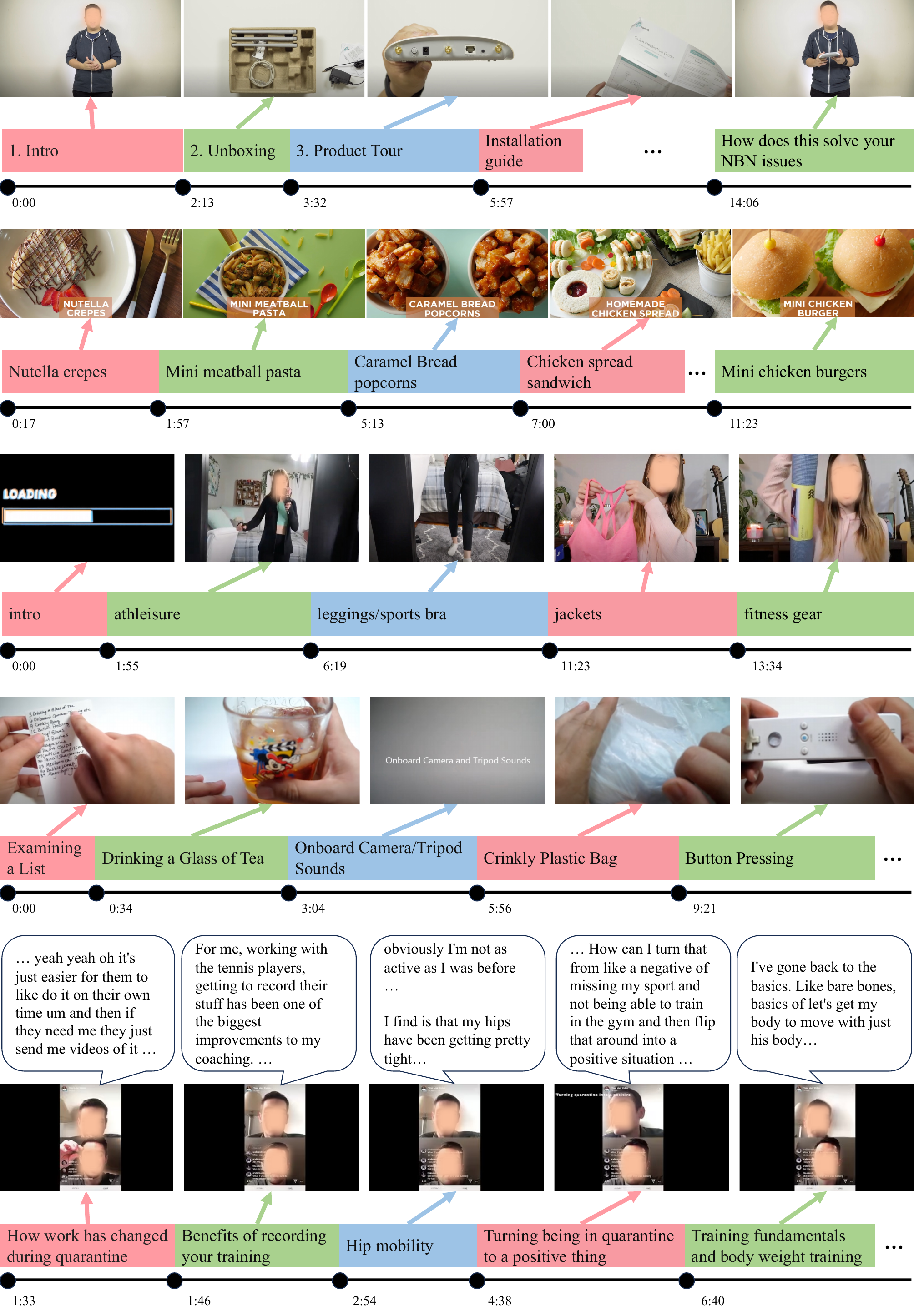}
\caption{
\textbf{Additional examples of videos with user-annotated chapters in \dataset{}:} 
Chapters depict visual events (e.g., the mini chicken burgers that appear in the second video), conversations (see the last video), or events in the raw audio (e.g., the sound of the crinkly plastic bag in the penultimate video) in various scenarios.
}
\label{fig:qualitative}
\end{figure*}

In Figure~\ref{fig:qualitative}, we provide additional examples that complement Figure\if\sepappendix1{ 1 of the main paper}\else{~\ref{fig:teaser}}\fi.
These examples illustrate the diversity of the data in \dataset{}, e.g., our dataset includes review videos, cooking videos, clothing fitting videos, ASMR videos, and videos of conversations.
These examples also show the multi-modal nature of the chapter data.
Indeed, chapters depict visual events (e.g., the mini chicken burgers that appear in the second video), conversations (see the last video), or events in the raw audio (e.g., the sound of the crinkly plastic bag in the penultimate video) in various scenarios.

\section{Qualitative examples of video chapter generation and dense video caption prediction}\label{supp:addpred}
\begin{figure}[t]
\centering
\includegraphics[width=1.\linewidth]{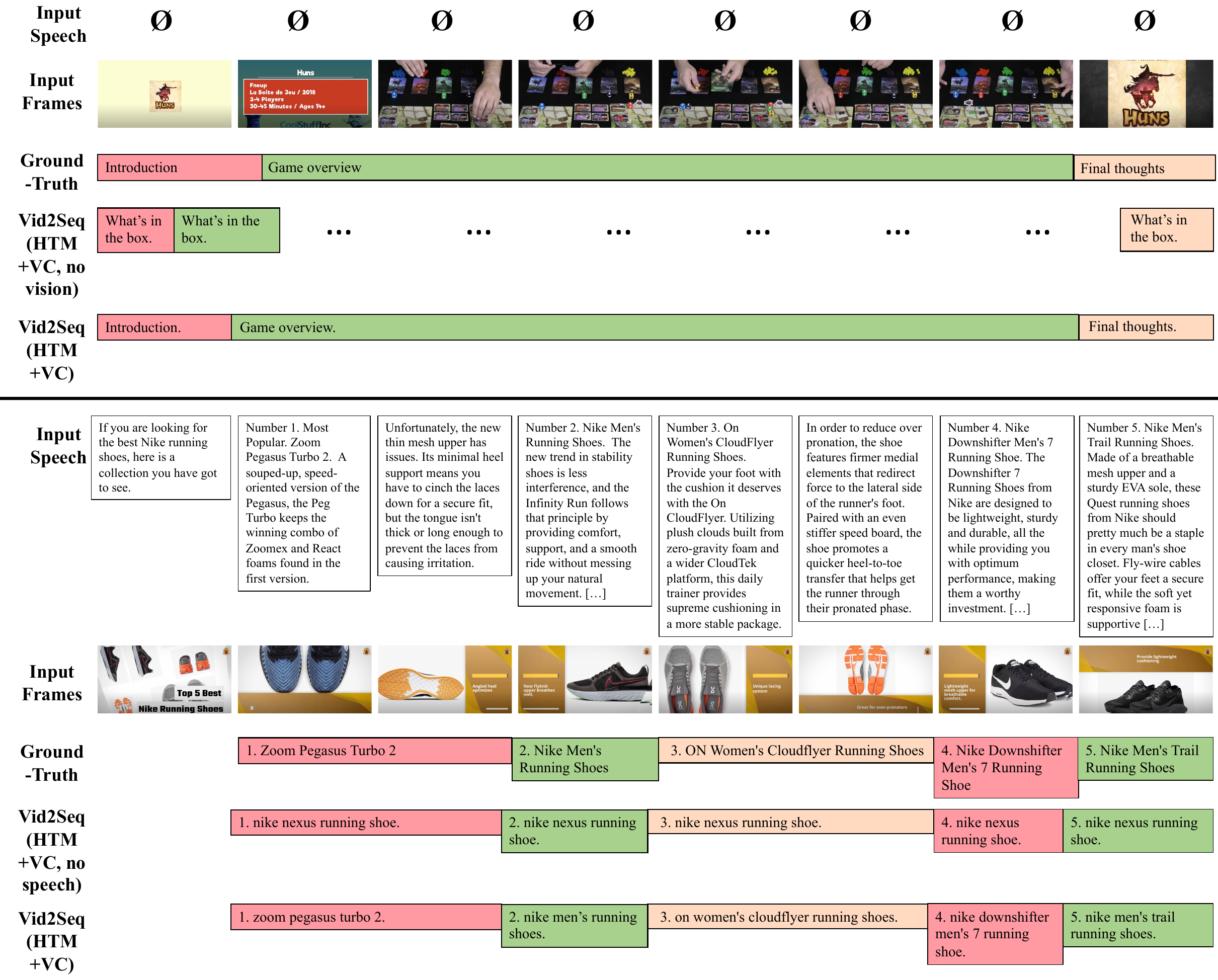}
\caption{
\textbf{Examples of video chapter generation using the Vid2Seq model with different input modalities compared with ground-truth on the test set of \dataset{}}.
The first example shows that the Vid2Seq variant with both speech and visual modalities "Vid2Seq (HTM+VC)" can predict the structure of the input video without the ASR input, unlike the Vid2Seq speech-only variant "Vid2Seq (HTM+VC, no vision)".
The second example shows that the Vid2Seq variant with both speech and visual modalities "Vid2Seq
(HTM +VC)" can effectively leverage speech cues to detect the names of the depicted and discussed shoes, unlike the Vid2Seq visual-only variant "Vid2Seq (HTM+VC, no
speech)".}
\label{fig:qualitativevcg}
\end{figure}

\begin{figure}[t]
\centering
\includegraphics[width=1.\linewidth]{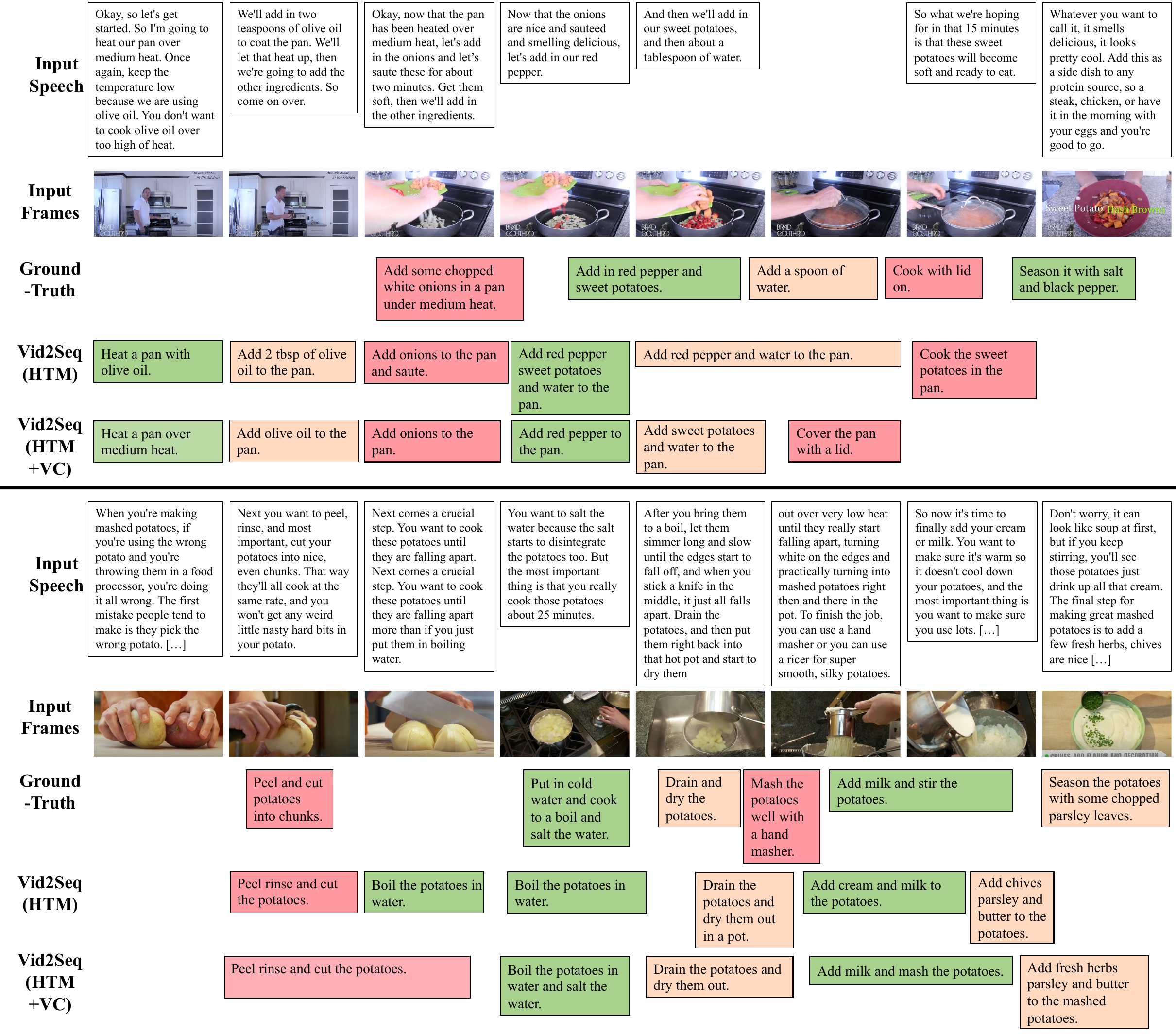}
\caption{
\textbf{Examples of dense event captioning of the Vid2Seq model pretrained on \dataset{} (vs. not pre-trained), compared with ground-truth, on the validation set of YouCook2.}
We find that the model pretrained on \dataset{} "Vid2Seq (HTM+VC)" is more accurate and less prone to hallucination.
For instance, in the first example (top), the non-VC-pretrained model "Vid2Seq (HTM)" predicts "Add red pepper sweet potatoes and water to the pan." before the sweet potatoes are actually thrown into the pan.
In the second example (bottom), the non-VC-pretrained model "Vid2Seq (HTM)" predicts the event "Boil the potatoes in water." twice and predicts the event "Add chives parsley and butter to the potatoes." before it actually happens. The VC-pretrained model "Vid2Seq (HTM+VC)" produces more accurate predictions.}
\label{fig:qualitativedvc}
\end{figure}

We present qualitative results for video chapter generation and dense video captioning in Figures~\ref{fig:qualitativevcg} and~\ref{fig:qualitativedvc}.
Compared with the speech-only model, a key advantage of the speech+visual video chapter generation model is that it can generalize to videos that do not have ASR input, as shown in the first example of Figure~\ref{fig:qualitativevcg}.
Compared with the visual-only variant, the multi-modal model can also benefit from speech cues, as seen in the second example in Figure~\ref{fig:qualitativevcg}.
Moreover, we observe that the dense video captioning model pretrained on \dataset{} is more accurate and hallucinates less than the variant not pretrained on \dataset{}, see Figure~\ref{fig:qualitativedvc}.

\section{Additional data analysis of \dataset{}}\label{supp:adddata}
\begin{figure}[!htbp]
\centering
\begin{subfigure}[valign=t]{.99\textwidth}
\includegraphics[width=\linewidth]{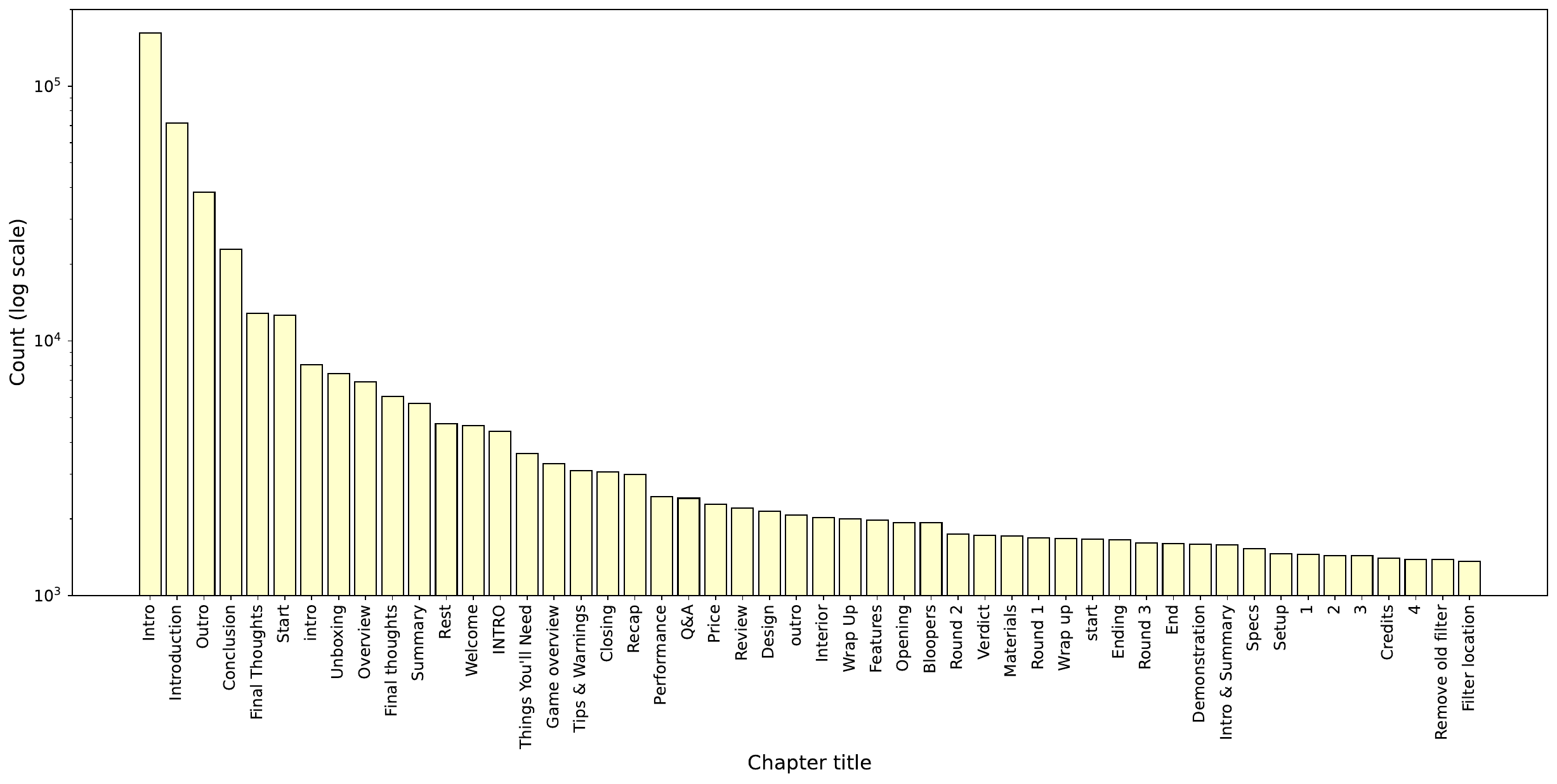}
\caption{Most common chapter titles}
\end{subfigure}
\begin{subfigure}{.99\textwidth}
\includegraphics[width=\linewidth]{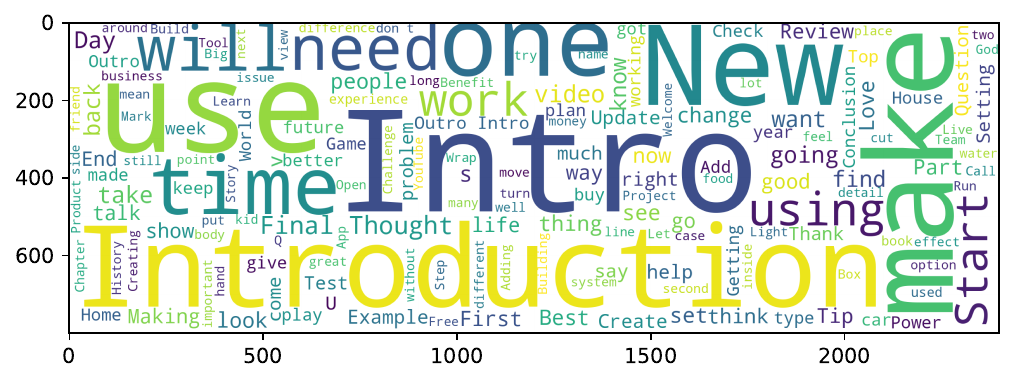}
\caption{Word clouds of chapter titles.}
\end{subfigure}
\begin{subfigure}{.99\textwidth}
\includegraphics[width=\linewidth]{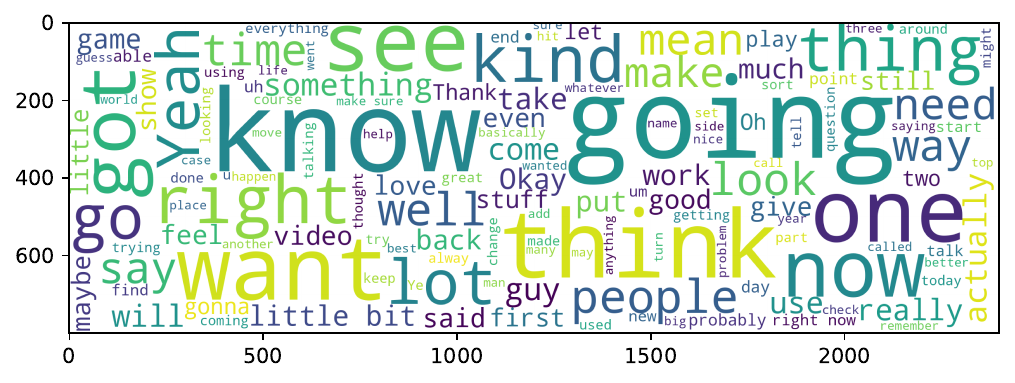}
\caption{Word clouds of ASR.}
\end{subfigure}
\caption{{\bf Additional statistics of the \dataset{} dataset.}}
\label{fig:addstats}
\end{figure}

\begin{figure}[t]
\centering
\includegraphics[width=.5\linewidth]{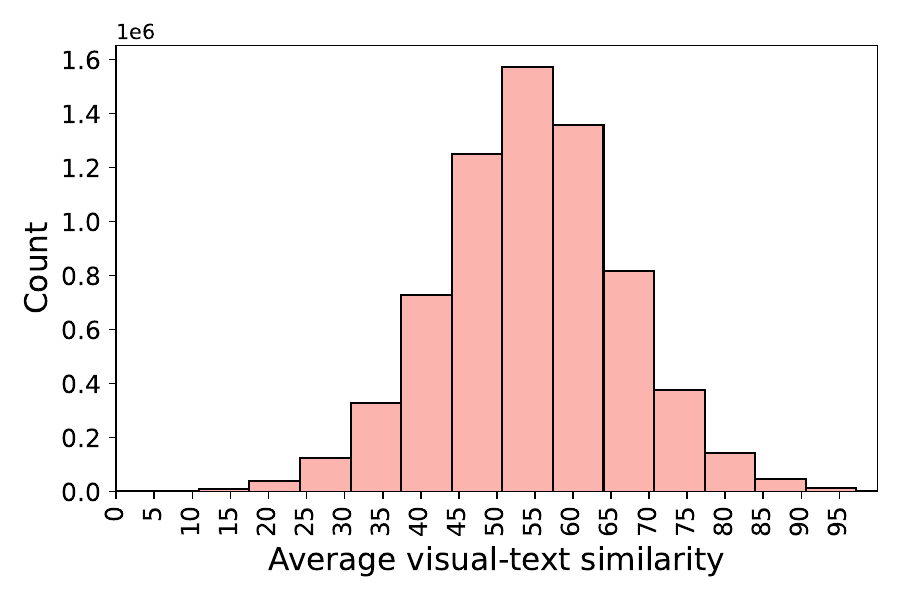}
\caption{\textbf{Average visual-text similarity between chapter titles and the corresponding video frames as measured by CLIP cosine similarity (rescaled between 0 and 100) in \dataset{}.}}
\label{fig:vistext}
\end{figure}

We here complement the analysis of the data in \dataset{} provided in Section\if\sepappendix1{ 3.3 of the main paper}\else{~\ref{sec:analysis}}\fi.
In Figure~\ref{fig:addstats}, we show a histogram of the most common chapter titles and word clouds\footnote{To generate the word clouds, we used \url{https://github.com/amueller/word_cloud}.} of the chapters titles and ASR content in \dataset{}. 
A few generic chapter titles that outline the structure of the video (e.g., \textit{Intro}, \textit{Introduction}, \textit{Outro}, \textit{Conclusion} and \textit{Start}) appear more than 10K times.
Besides, we notice that many videos include chapters about \textit{Unboxing}, \textit{Review}, or \textit{Tips}.
This is consistent with the fact that there are many vlogs and 'Howto' videos in \dataset{}.
We also observe that the most common words in the ASR largely differ from the most common words in the chapter titles, which further shows the difference between these two types of data.
To further measure the text-video alignment in the \dataset{} dataset, we compute the CLIP cosine similarity between chapter titles and their corresponding video frames and plot the resulting distribution in Figure~\ref{fig:vistext}.
The average similarity score is 54.6\%, and less than 1\% of the chapters have a visual-text similarity score below 30\%. These statistics demonstrate a good video-text alignment in the VidChapters-7M dataset.

\section{Additional implementation details}\label{supp:addimplem}
In this Section, we present implementation details that complement the information provided in Section\if\sepappendix1{ 4 of the main paper}\else{~\ref{sec:experiments}}\fi.
We discuss implementation details of our tagging protocol for ethical considerations in Section~\ref{sec:detailsprocessing}, models used for video chapter generation and dense video captioning in Section~\ref{sec:detailsvcgdvc}, models used for video chapter generation with ground-truth boundaries in Section~\ref{sec:detailsvcggt}, and models used for video chapter grounding in Section~\ref{sec:detailsvcgr}.

\subsection{Tagging for ethical considerations}\label{sec:detailsprocessing}
In Section\if\sepappendix1{ 3.3 of the main paper}\else{~\ref{sec:analysis}}\fi, we explained how we tag videos for ethical considerations.
We here give additional details about this procedure.
For the NSFW visual content detector~\cite{schuhmannlaion}, we compute the NSFW score at every frame (at 1 FPS) and tag videos with an average score above 0.5.
For the toxic content detection model~\cite{Detoxify}, we compute the toxicity score at every chapter title / ASR sentence and tag videos where the chapter titles / ASR have an average toxicity score above 0.5.

\subsection{Video chapter generation and dense video captioning}\label{sec:detailsvcgdvc}

\noindent \textbf{LLaMA~\cite{touvron2023llama}.} 
We use the following prompt: \texttt{Summarize the following speech transcript in a chapter title. Transcript: <ASR> Chapter title:} where the ASR is the concatenation of all speech sentences spoken during a given video segment.

\noindent \textbf{BLIP-2~\cite{li2023blip}.} 
We use the following prompt: \texttt{Summarize the image in a chapter title. Chapter title:}, and use the middle frame of the predicted video segment.

\noindent \textbf{PDVC~\cite{wang2021end}.}
We use PDVC's official codebase.
PDVC includes a caption decoder that relies on dataset-specific word vocabularies.
To adapt PDVC to \dataset{}/YouCook2/ViTT, we construct a vocabulary made with all words that appear at least 50/2/3 times in the dataset (33,598/3,815/1,607 words).
For transfer learning from \dataset{} to YouCook2/ViTT, we initialize the downstream dataset-specific word embedding layer with the weights of the corresponding word embedding in the pretrained model.
We subsample or pad the sequence of frames to 100 frames.
For all datasets, we use 100 queries and train with a constant learning rate of $5e^{-5}$, weight decay $1e^{-4}$ and batch size 1 on an NVIDIA V100 32GB (as the official codebase is not compatible with higher batch sizes or multi-gpu training) .
We train on \dataset{}/YouCook2/ViTT for 5/30/30 epochs.
The training on \dataset{} lasts about a week.

\noindent \textbf{Vid2Seq~\cite{yang2023vid2seq}.}
We reimplement Vid2Seq (originally released in Jax) in PyTorch.
For initialization, we use the T5-Base language model pretrained on the C4 text corpus~\cite{raffel2020exploring}.
Vid2Seq is originally pretrained on YT-Temporal-1B~\cite{zellers2022merlot} using a generative and denoising objective in the speech sequence.
Due to computational limitations, we instead pretrain Vid2Seq on the smaller HowTo100M dataset~\cite{miech19howto100m} with the same objectives. 
Then we train Vid2Seq on \dataset{} with the next token prediction objective in the chapter sequence and the denoising objective in the speech sequence.
Finetuning on YouCook2/ViTT is done with the next token prediction objective in the dense video captioning sequence and the denoising objective in the speech sequence.
We subsample or zero-pad the sequence of frames to 100 frames.
The text encoder and decoder sequence are truncated or padded to 1000 and 256 tokens, respectively.
For all datasets, we use a learning rate of $3e^{-4}$ warmed up linearly (from 0) for the first 10\% of iterations and following a cosine decay (down to 0) for the remaining 90\%.
We train for 6/10/40/20 epochs on HowTo100M/\dataset{}/YouCook2/ViTT. 
We use a batch size of 64 videos split on 8 NVIDIA A100 80GB for HowTo100M/\dataset{}, and 16 videos split on 8 NVIDIA V100 32GB for YouCook2/ViTT.
The training on HowTo100M or \dataset{} takes about 2 days.

\subsection{Video chapter generation with ground-truth boundaries}\label{sec:detailsvcggt}

\noindent \textbf{LLaMA~\cite{touvron2023llama} and BLIP-2~\cite{li2023blip}.} See Section~\ref{sec:detailsvcgdvc}.

\noindent \textbf{Vid2Seq~\cite{yang2023vid2seq}.} 
To adapt the model pretrained on HowTo100M (see Section~\ref{sec:detailsvcgdvc}) to video chapter generation with ground-truth boundaries, we remove the model weights corresponding to the time tokens (in the token embedding layers and the token prediction layer).
We train for 20 epochs on \dataset{} using the next token prediction objective in the sequence of tokens corresponding to a single chapter title. 
We construct training batches by sampling a chapter title with its associated video clip at each iteration (i.e., an epoch corresponds to seeing one chapter title for all videos).
The text encoder and decoder sequence are truncated or padded to 256 and 32 tokens, respectively.
We use a learning rate of $3e^{-4}$ warmed up linearly (from 0) for the first 10\% of iterations and following a cosine decay (down to 0) for the remaining 90\%.
We use a batch size of 512 videos split on 8 NVIDIA A100 80GB for \dataset{}.
The training takes about a day.

\subsection{Video chapter grounding}\label{sec:detailsvcgr}

\noindent \textbf{Moment-DETR~\cite{lei2021detecting}.}
We use Moment-DETR's official codebase.
We train with the AdamW optimizer~\cite{loshchilov2017decoupled}, a constant learning rate of $3e^{-4}$, and a batch size of 256 videos split on 8 NVIDIA A100 80GB.
We use a FPS of 1/3 and subsample or zero-pad the sequence of frames to 1200 frames.
We use a maximum number of text query tokens of 77.
We train for 50 epochs on \dataset{}, where an epoch corresponds to seeing one chapter title for all videos, which takes about 2 days.

\section{Video chapter generation results split by language}\label{supp:addres}
\begin{table}[t]
\begin{center}
\setlength\tabcolsep{3pt}
\resizebox{1.\linewidth}{!}{
\begin{tabular}{l|ccc|cccccccc}
Method & Modalities & Pretraining Data & Finetuned & S & B1 & B2 & B3 & B4 & C & M & RL \\
\hline
Text tiling~\cite{hearst1997text} + Random & Speech & $\emptyset$ & \xmark
& 0.5 & 0.8 & 0.2 & 0.1 & 0.0 & 0.9 & 0.8 & 0.7 \\
Text tiling~\cite{hearst1997text} + LLaMA~\cite{touvron2023llama} & Speech & Text mixture & \xmark
& 0.3 & 0.5 & 0.2 & 0.1 & 0.0 & 0.5 & 0.4 & 0.4 \\
Shot detect~\cite{tomar2006converting} + BLIP-2~\cite{li2023blip} & Visual & 129M image-texts & \xmark
& 1.3 & 1.5 & 0.7 & 0.4 & 0.2 & 4.7 & 1.4 & 1.6 \\
\hline
PDVC~\cite{wang2021end} & Visual & $\emptyset$ & \cmark
& 6.6 & 9.0 & 3.8 & 1.5 & 1.0 & 36.0 & 9.1 & 11.0 \\
Vid2Seq~\cite{yang2023vid2seq} & Speech+Visual & C4 & \cmark
& 10.8 & 10.3 & 7.6 & 4.9 & 3.4 & 54.8 & 9.1 & 11.9 \\
Vid2Seq~\cite{yang2023vid2seq} w/ mT5 & Speech+Visual & mC4 & \cmark
& 10.4 & 9.9 & 7.2 & 4.7 & 3.3 & 52.0 & 8.7 & 11.3 \\
Vid2Seq~\cite{yang2023vid2seq} & Speech+Visual & C4 + HowTo100M & \cmark
& \textbf{11.5} & \textbf{11.1} & \textbf{8.1} & \textbf{5.1} & \textbf{3.6} & \textbf{58.8} & \textbf{9.7} & \textbf{12.8} \\
\end{tabular}
}
\vspace{0.1cm}
\caption{\textbf{Video chapter generation (global metrics) on the \dataset{} test set restricted to videos with English chapter titles and ASR.}
Here, finetuned refers to finetuning on the \dataset{} train set, and speech refers to transcribed speech (ASR).}
\label{table:vcgen}
\end{center}
\end{table}

\begin{table}[t]
\begin{center}
\setlength\tabcolsep{3pt}
\resizebox{1.\linewidth}{!}{
\begin{tabular}{l|ccc|cccccccc}
Method & Modalities & Pretraining Data & Finetuned & S & B1 & B2 & B3 & B4 & C & M & RL \\
\hline
Text tiling~\cite{hearst1997text} + Random & Speech & $\emptyset$ & \xmark
& 0.6 & 1.7 & 1.3 & 1.3 & 1.1 & 12.8 & 1.5 & 1.6 \\
Text tiling~\cite{hearst1997text} + LLaMA~\cite{touvron2023llama} & Speech & Text mixture & \xmark
& 0.1 & 0.3 & 0.2 & 0.0 & 0.0 & 0.0 & 0.2 & 0.2 \\
Shot detect~\cite{tomar2006converting} + BLIP-2~\cite{li2023blip} & Visual & 129M image-texts & \xmark
& 0.6 & 0.4 & 0.2 & 0.0 & 0.0 & 1.3 & 0.6 & 0.5 \\
\hline
PDVC~\cite{wang2021end} & Visual & $\emptyset$ & \cmark
& 5.4 & \textbf{11.6} & 0.0 & 0.0 & 0.0 & 29.4 & \textbf{12.4} & \textbf{14.9} \\
Vid2Seq~\cite{yang2023vid2seq} & Speech+Visual & C4 & \cmark
& 9.1 & 8.4 & 5.2 & 1.0 & 0.9 & 34.1 & 6.1 & 10.1 \\
Vid2Seq~\cite{yang2023vid2seq} w/ mT5 & Speech+Visual & mC4 & \cmark
& 8.8 & 8.1 & 5.9 & 1.7 & 1.8 & 38.4 & 6.1 & 10.1 \\
Vid2Seq~\cite{yang2023vid2seq} & Speech+Visual & C4 + HowTo100M & \cmark
& \textbf{10.9} & 9.6 & \textbf{5.4} & \textbf{1.7} & \textbf{1.7} & \textbf{43.2} & 8.1 & 8.1 \\
\end{tabular}}
\vspace{0.1cm}
\caption{\textbf{Video chapter generation (global metrics) on the \dataset{} test set restricted to videos with German chapter titles and ASR.}
Here, finetuned refers to finetuning on the \dataset{} train set, and speech refers to transcribed speech (ASR).}
\label{table:vcgde}
\end{center}
\end{table}

We report video chapter generation results on the \dataset{} dataset split by language for both English and German in Tables~\ref{table:vcgen} and~\ref{table:vcgde}, respectively. 
We find that training on \dataset{} is beneficial for both languages. 
Interestingly, pretraining on HowTo100M (which is a dataset in English) improves results on English as well as German. 
We also observe that the quantitative results in German are lower than in English. 
Finally, we report results of the Vid2Seq model with the multi-lingual language model mT5~\cite{xue2020mt5} pretrained on the multi-lingual dataset mC4~\cite{xue2020mt5}. 
We find that this variant performs a bit worse on English but slightly better on German compared to the Vid2Seq variant based on T5 pretrained on the C4 corpus.

\section{Datasheet for \dataset{}}\label{supp:datasheet}
Datasheets for datasets introduced by \citet{gebru2021datasheets} serve as a medium of communication between the creators and users of a dataset.
They effectively consolidate the motivation, creation process, composition, and intended uses of a dataset as a series of questions and answers.
In this Section, we provide a datasheet for the \dataset{} dataset.

\section*{Motivation}

\begin{compactenum}[\hspace{0pt}Q1.]
\setcounter{enumi}{0}

\dsquestionex{For what purpose was the dataset created?}{Was there a specific task in mind? Was there a specific gap that needed to be filled? Please provide a description.}
\label{Q1}

\dsanswer{
The \dataset{} dataset was created to explore the task of video chapter generation, which enables users to quickly navigate to the information of their interest.
}

\dsquestion{Who created this dataset (e.g., which team, research group) and on behalf of which entity (e.g., company, institution, organization)?}
\label{Q2}

\dsanswer{
Five researchers have created \dataset{}: Antoine Yang (Inria and DI ENS), Arsha Nagrani (VGG, University of Oxford), Ivan Laptev (Inria and DI ENS), Josef Sivic (CIIRC CTU) and Cordelia Schmid (Inria and DI ENS).
}

\dsquestionex{Who funded the creation of the dataset?}{If there is an associated grant, please provide the name of the grantor and the grant name and number.}
\label{Q3}

\dsanswer{
We collected \dataset{} without any monetary costs, since no part of our dataset requires annotations from crowd workers or contractors.
This research work was funded by Antoine Yang's Google PhD fellowship, the French government under management of Agence Nationale de la Recherche as part of the "Investissements d'avenir" program, reference ANR-19-P3IA-0001 (PRAIRIE 3IA Institute), the Louis Vuitton ENS Chair on Artificial Intelligence, the European Regional Development Fund under project IMPACT (reg.\ no.\ CZ.02.1.01/0.0/0.0/15 003/0000468).
However, note that this article solely reflects the opinions and conclusions of its authors and not of its funders.
}

\dsquestion{Any other comments?}
\label{Q4}

\dsanswer{No.}

\end{compactenum}

\section*{Composition}

\begin{compactenum}[\hspace{0pt}Q1.]
\setcounter{enumi}{4}
    
\dsquestionex{What do the instances that comprise the dataset represent (e.g., documents, photos, people, countries)?}{ Are there multiple types of instances (e.g., movies, users, and ratings; people and interactions between them; nodes and edges)? Please provide a description.}
\label{Q5}

\dsanswer{
Each instance in \dataset{} represents a YouTube video.
}

\dsquestion{How many instances are there in total (of each type, if appropriate)?}
\label{Q6}

\dsanswer{
There are \dsetsize{} instances in \dataset{}.
}

\dsquestionex{Does the dataset contain all possible instances or is it a sample (not necessarily random) of instances from a larger set?}{ If the dataset is a sample, then what is the larger set? Is the sample representative of the larger set (e.g., geographic coverage)? If so, please describe how this representativeness was validated/verified. If it is not representative of the larger set, please describe why not (e.g., to cover a more diverse range of instances, because instances were withheld or unavailable).}
\label{Q7}

\dsanswer{
\dataset{} is a small sample drawn from all the data uploaded to YouTube. 
Millions of videos are uploaded on YouTube every day.
We started from a subset of 92 million YouTube video candidates, which consists of videos recommended in 
videos from the YT-Temporal-180M dataset~\cite{zellers2021merlot}.
We selected the videos from this subset (\dsetsize{} instances) that contain user-annotated chapters.
Hence, \dataset{} data does not fully represent YouTube.
}

\dsquestionex{What data does each instance consist of? “Raw” data (e.g., unprocessed text or images) or features?}{In either case, please provide a description.}
\label{Q8}

\dsanswer{
Each instance in \dataset{} consists of four metadata fields:
\begin{compactitem}
\item \texttt{"video\_id"}: Unique alphanumeric ID of the video (assigned by YouTube).
\item \texttt{"url"}: Static URL for downloading the video, e.g., \texttt{https://www.youtube.com/watch?v=<video\_id>}.
\item \texttt{"asr"}: ASR transcripts aligned over time.
\item \texttt{"chapters"}: Chapters aligned over time.
\end{compactitem}
}

\dsquestionex{Is there a label or target associated with each instance?}{If so, please provide a description.}
\label{Q9}

\dsanswer{
We use the chapters as labels in this work, though it might be also possible to use auxiliary
information (like video titles or tags).
}

\dsquestionex{Is any information missing from individual instances?}{If so, please provide a description, explaining why this information is missing (e.g., because it was unavailable). This does not include intentionally removed information, but might include, e.g., redacted text.}
\label{Q10}

\dsanswer{
No and yes.
No, because all the metadata fields for every instance are filled with valid values.
Yes, because the \texttt{"url"} for some instances may not retrieve the underlying video.
This may happen if the YouTube user (author) removes the video from YouTube.
Such deletions reduce our dataset size over time, however, video deletions are rare.
}

\dsquestionex{Are relationships between individual instances made explicit (e.g., users’ movie ratings, social network links)?}{If so, please describe how these relationships are made explicit.}
\label{Q11}

\dsanswer{
Relationships between individual instances (e.g., videos made by the same creator) are not made explicit in our work, though this is a possibility for future work.
}

\dsquestionex{Are there recommended data splits (e.g., training, development/validation, testing)?}{If so, please provide a description of these splits, explaining the rationale behind them.}
\label{Q12}

\dsanswer{
We randomly split our data into training, validation, and testing sets.
The training, validation, and testing sets are meant for training, development, and evaluation, respectively.
}

\dsquestionex{Are there any errors, sources of noise, or redundancies in the dataset?}{If so, please provide a description.}
\label{Q13}

\dsanswer{
\dataset{} is inherently noisy since YouTubers are free to write the chapters that they want.
}

\dsquestionex{Is the dataset self-contained, or does it link to or otherwise rely on external resources (e.g., websites, tweets, other datasets)?}{If it links to or relies on external resources,
\begin{compactenum}[\hspace{1pt}(a)]
    \item Are there guarantees that they will exist, and remain constant, over time?
    \item Are there official archival versions of the complete dataset (i.e., including the external resources as they existed at the time the dataset was created)?
    \item Are there any restrictions (e.g., licenses, fees) associated with any of the external resources that might apply to a future user? Please provide descriptions of all external resources and any restrictions associated with them, as well as links or other access points, as appropriate.
\end{compactenum}}
\label{Q14}

\dsanswer{
We do not distribute videos of our dataset to respect YouTube user privacy and to limit our storage budget.
Instead, we provide video URLs (\texttt{"url"}, \qref{Q8}) that point to videos hosted on YouTube servers.
In response to sub-questions:
\begin{compactenum}[\hspace{1pt}(a)]
    \item These video servers ensure stable access unless the YouTube user deletes their video.
    \item Yes, YouTube archives all the metadata of submitted videos. For videos, YouTube only archives the URL and not the media content, giving full control of accessibility to the users.
    \item All video URLs are freely accessible. It is unlikely for video servers to restrict access in the future, given their free accessibility over the past decade.
\end{compactenum}
}

\dsquestionex{Does the dataset contain data that might be considered confidential (e.g., data that is protected by legal privilege or by doctor-patient confidentiality, data that includes the content of individuals non-public communications)?}{If so, please provide a description.}
\label{Q15}

\dsanswer{
No, the videos included in \dataset{} do not cover topics that may be considered confidential.
All videos were publicly shared on YouTube prior to inclusion in \dataset{}.
}

\dsquestionex{Does the dataset contain data that, if viewed directly, might be offensive, insulting, threatening, or might otherwise cause anxiety?}{If so, please describe why.}
\label{Q16}

\dsanswer{
The scale of \dataset{} means that we are unable to manually verify the contents of all videos and chapters.
However, YouTube removes videos that contain offensive content or do not follow their community guidelines.
Furthermore, we employed additional mitigation techniques on \dataset{}:
\begin{compactenum}
    \item We tagged all instances whose video frames were predicted as NSFW by an off-the-shelf detector~\cite{schuhmannlaion}.
    \item We tagged all instances whose chapter titles or speech transcripts were predicted as toxic by a language model~\cite{Detoxify}.
\end{compactenum}
}

\dsquestionex{Does the dataset relate to people?}{If not, you may skip remaining questions in this section.}
\label{Q17}

\dsanswer{
The dataset pertains to people in that people upload videos to YouTube and write descriptions that include chapter annotations.
Furthermore, most videos in \dataset{} have people speaking and/or appearing.
}

\dsquestionex{Does the dataset identify any subpopulations (e.g., by age, gender)?}{If so, please describe how these subpopulations are identified and provide a description of their respective distributions within the dataset.}
\label{Q18}

\dsanswer{
\dataset{} does not explicitly identify any subpopulations.
Since most videos contain people and chapters are free-form natural language written by YouTube users, it is possible that some chapters may identify people appearing in individual videos as part of a subpopulation.
}

\dsquestionex{Is it possible to identify one or more natural persons, either directly or indirectly (i.e., in combination with other data) from the dataset?}{If so, please describe how.}
\label{Q19}

\dsanswer{
Yes, our data includes celebrities, or other YouTube-famous people. 
All of the videos that we use are of publicly available data, following the Terms of Service (\url{https://www.youtube.com/static?template=terms}) that users agreed to when uploading to YouTube.
}

\dsquestionex{Does the dataset contain data that might be considered sensitive in any way (e.g., data that reveals racial or ethnic origins, sexual orientations, religious beliefs, political opinions or union memberships, or locations; financial or health data; biometric or genetic data; forms of government identification, such as social security numbers; criminal history)?}{If so, please provide a description.}
\label{Q20}

\dsanswer{
This is highly unlikely, as YouTube removes videos that contain offensive content or do not follow their community guidelines.
}

\dsquestion{Any other comments?}
\label{Q21}

\dsanswer{No.}

\end{compactenum}

\section*{Collection Process}

\begin{compactenum}[\hspace{0pt}Q1.]
\setcounter{enumi}{21}

\dsquestionex{How was the data associated with each instance acquired?}{Was the data directly observable (e.g., raw text, movie ratings), reported by subjects (e.g., survey responses), or indirectly inferred/derived from other data (e.g., part-of-speech tags, model-based guesses for age or language)? If data was reported by subjects or indirectly inferred/derived from other data, was the data validated/verified? If so, please describe how.}
\label{Q22}

\dsanswer{
See \qref{Q7} for an explanation of how the candidate video IDs were chosen.
These video IDs were provided by the YT-Temporal-180M dataset providers~\cite{zellers2021merlot} and collected via the YouTube API. 
The \texttt{"video\_id"} and \texttt{"URL"} are directly observable. 
The \texttt{"chapters"} are extracted from the YouTube description which is directly observable.
The \texttt{"asr"} is obtained by applying the Whisper-Large-V2 model~\cite{radford2022robust} to the directly observable audio from the video. 
We found this model to provide higher-quality transcriptions compared to the YouTube API on several data samples from \dataset{}.
}

\dsquestionex{What mechanisms or procedures were used to collect the data (e.g., hardware apparatus or sensor, manual human curation, software program, software API)?}{How were these mechanisms or procedures validated?}
\label{Q23}

\dsanswer{
We collected all data using compute resources provided by IDRIS, under the allocation 2023-A0131011670 made by GENCI.
The code for querying APIs, extracting ASR, and filtering data are implemented in Python.
The code was validated by checking several data samples from \dataset{}.
}

\dsquestion{If the dataset is a sample from a larger set, what was the sampling strategy?}
\label{Q24}

\dsanswer{
See \qref{Q7}.
}

\dsquestion{Who was involved in data collection process (e.g., students, crowd-workers, contractors) and how were they compensated (e.g., how much were crowd-workers paid)?}
\label{Q25}

\dsanswer{
Our data collection pipeline is fully automatic and does not require any human annotators.
YouTube users have uploaded videos whose metadata is a part of \dataset{} -- we did not directly interact with these users.
}

\dsquestionex{Over what timeframe was the data collected? Does this timeframe match the creation timeframe of the data associated with the instances (e.g., recent crawl of old news articles)?}{If not, please provide a description of the timeframe.}
\label{Q26}

\dsanswer{
\dataset{} contains videos that were uploaded to YouTube between 2005--2022.
We collected all data in early 2023, which we used to conduct experiments for our NeurIPS 2023 submission.
}

\dsquestionex{Were any ethical review processes conducted (e.g., by an institutional review board)?}{If so, please provide a description of these review processes, including the outcomes, as well as a link or other access point to any supporting documentation.}
\label{Q27}

\dsanswer{
We did not conduct a formal ethical review process via institutional review boards.
However, as described in \if\sepappendix1{\textcolor{red}{Section 3.3} of the main paper }\else{\textcolor{red}{Section~\ref{sec:analysis} }}\fi and \qref{Q16} we employed several filtering mechanisms to tag instances that could be problematic.
}

\dsquestionex{Does the dataset relate to people?}{If not, you may skip remaining questions in this section.}
\label{Q28}

\dsanswer{
Yes, see \qref{Q17}.
}

\dsquestion{Did you collect the data from the individuals in question directly, or obtain it via third parties or other sources (e.g., websites)?}
\label{Q29}

\dsanswer{
We collected data submitted by YouTube users indirectly through the YouTube API.
However, users agree with YouTube's Terms of Service regarding the redistribution of their data by YouTube.
}

\dsquestionex{Were the individuals in question notified about the data collection?}{If so, please describe (or show with screenshots or other information) how notice was provided, and provide a link or other access point to, or otherwise reproduce, the exact language of the notification itself.}
\label{Q30}

\dsanswer{
No. YouTube users are not required to share their personal contact information (email, phone numbers, etc.).
Hence, the only way to notify the authors of \dataset{} videos is by commenting on their videos.
This is practically difficult to do manually and will be classified as spam and blocked by YouTube if attempted to programmatically write a templated comment to millions of users.
}

\dsquestionex{Did the individuals in question consent to the collection and use of their data?}{If so, please describe (or show with screenshots or other information) how consent was requested and provided, and provide a link or other access point to, or otherwise reproduce, the exact language to which the individuals consented.}
\label{Q31}

\dsanswer{
Users did not explicitly consent to the use of their data in our dataset.
However, by uploading their data on YouTube, they consent that it would appear on the YouTube plaform and will be accessible via the official YouTube API (which we use to collect \dataset{}).
}

\dsquestionex{If consent was obtained, were the consenting individuals provided with a mechanism to revoke their consent in the future or for certain uses?}{If so, please provide a description, as well as a link or other access point to the mechanism (if appropriate).}
\label{Q32}

\dsanswer{
Users have full control over the presence of their data in our dataset.
If users wish to revoke their consent, they can delete the underlying YouTube video -- it will be automatically removed from \dataset{} since we distributed videos as URLs.
Moreover, we provide an opt-out request form on our dataset website for anybody to request removal of an individual instance if it is potentially harmful (e.g. NSFW, violates privacy, harmful stereotypes, etc.).
}

\dsquestionex{Has an analysis of the potential impact of the dataset and its use on data subjects (e.g., a data protection impact analysis) been conducted?}{If so, please provide a description of this analysis, including the outcomes, as well as a link or other access point to any supporting documentation.}
\label{Q33}

\dsanswer{No.}

\dsquestion{Any other comments?}
\label{Q34}

\dsanswer{No.}

\end{compactenum}

\section*{Preprocessing, Cleaning, and/or Labeling}

\begin{compactenum}[\hspace{0pt}Q1.]
\setcounter{enumi}{34}

\dsquestionex{Was any preprocessing/cleaning/labeling of the data done (e.g., discretization or bucketing, tokenization, part-of-speech tagging, SIFT feature extraction, removal of instances, processing of missing values)?}{If so, please provide a description. If not, you may skip the remainder of the questions in this section.}
\label{Q35}

\dsanswer{
We converted chapter timestamps in HH:MM:SS format to seconds.
Refer to \if\sepappendix1{\textcolor{red}{Section 3.1} of the main paper }\else{\textcolor{red}{Section~\ref{sec:collection} }}\fi for more details.
We also extracted speech transcripts and visual features (see \if\sepappendix1{\textcolor{red}{Section 3.2} of the main paper}\else{\textcolor{red}{Section~\ref{sec:processing}}}\fi).
Finally, we tagged some instances with a focus on ethical considerations, see \qref{Q16} for more details.
}

\dsquestionex{Was the ``raw'' data saved in addition to the preprocessed/cleaned/labeled data (e.g., to support unanticipated future uses)?}{If so, please provide a link or other access point to the “raw” data.}
\label{Q36}

\dsanswer{
Yes, the raw descriptions from which chapters are extracted are also released on the dataset website~\cite{chapterswebpage}.
}

\dsquestionex{Is the software used to preprocess/clean/label the instances available?}{If so, please provide a link or other access point.}
\label{Q37}

\dsanswer{
Yes, the data preprocessing code is open-sourced and accessible from the dataset website~\cite{chapterswebpage}.
}

\dsquestion{Any other comments?}
\label{Q38}

\dsanswer{No.}

\end{compactenum}

\section*{Uses}

\begin{compactenum}[\hspace{0pt}Q1.]
\setcounter{enumi}{38}

\dsquestionex{Has the dataset been used for any tasks already?}{If so, please provide a description.}
\label{Q39}

\dsanswer{
We have used our dataset to train deep neural networks that perform video chapter generation, and that can be transferred to dense video captioning tasks (see \if\sepappendix1{\textcolor{red}{Sections 4.1} and \textcolor{red}{4.4} in the main paper}\else{\textcolor{red}{Sections~\ref{sec:vcg}} and\textcolor{red}{~\ref{sec:dvc}}}\fi). 
We also trained models for video chapter generation with ground-truth boundaries and video chapter grounding (see \if\sepappendix1{\textcolor{red}{Sections 4.2} and \textcolor{red}{4.3} in the main paper}\else{\textcolor{red}{Sections~\ref{sec:vcggt}} and\textcolor{red}{~\ref{sec:vcgr}}}\fi).
}

\dsquestionex{Is there a repository that links to any or all papers or systems that use the dataset?}{If so, please provide a link or other access point.}
\label{Q40}

\dsanswer{
We do not maintain such a repository.
However, citation trackers like Google Scholar and Semantic Scholar would list all future works that cite our dataset.
}

\dsquestion{What (other) tasks could the dataset be used for?}
\label{Q41}

\dsanswer{
We anticipate that the dataset could be used for a variety of video-and-language tasks, such as text-to-video retrieval.
}

\dsquestionex{Is there anything about the composition of the dataset or the way it was collected and preprocessed/cleaned/labeled that might impact future uses?}{For example, is there anything that a future user might need to know to avoid uses that could result in unfair treatment of individuals or groups (e.g., stereotyping, quality of service issues) or other undesirable harms (e.g., financial harms, legal risks) If so, please provide a description. Is there anything a future user could do to mitigate these undesirable harms?}
\label{Q42}

\dsanswer{
This is very difficult to anticipate.
Future users of our dataset should be aware of YouTube's user demographics 
which might subtly influence the types of videos, languages, and ideas that are present in the dataset.
Also, note that our dataset is mainly composed of English videos, hence models trained on this dataset might perform worse on videos in other languages.
}

\dsquestionex{Are there any tasks for which the dataset should not be used?}{If so, please provide a description.}
\label{Q43}

\dsanswer{
Broadly speaking, our dataset should only be used for non-commercial academic research.
Our dataset should not be used for any tasks that involve identifying features related to people (facial recognition, gender, age, ethnicity identification, etc.) or making decisions that impact people (mortgages, job applications, criminal sentences; or moderation decisions about user-uploaded data that could result in bans from a website).
Any commercial and for-profit uses of our dataset are restricted -- it should not be used to train models that will be deployed in production systems as part of a product offered by businesses or government agencies.
}

\dsquestion{Any other comments?}
\label{Q44}

\dsanswer{No.}

\end{compactenum}

\section*{Distribution}

\begin{compactenum}[\hspace{0pt}Q1.]
\setcounter{enumi}{44}

\dsquestionex{Will the dataset be distributed to third parties outside of the entity (e.g., company, institution, organization) on behalf of which the dataset was created?}{If so, please provide a description.}
\label{Q45}

\dsanswer{
Yes, our dataset is publicly available.
}

\dsquestionex{How will the dataset will be distributed (e.g., tarball on website, API, GitHub)}{Does the dataset have a digital object identifier (DOI)?}
\label{Q46}

\dsanswer{
We distribute our dataset as JSON/PICKLE files containing annotations.
Users will have to download the videos by themselves by using our data collection code.
All uses of \dataset{} should cite the paper as the reference.
}

\dsquestion{When will the dataset be distributed?}
\label{Q47}

\dsanswer{
The dataset is publicly available as of September 2023.
}

\dsquestionex{Will the dataset be distributed under a copyright or other intellectual property (IP) license, and/or under applicable terms of use (ToU)?}{If so, please describe this license and/or ToU, and provide a link or other access point to, or otherwise reproduce, any relevant licensing terms or ToU, as well as any fees associated with these restrictions.}
\label{Q48}

\dsanswer{
Uses of our dataset are subject to YouTube API terms (\url{https://www.youtube.com/static?template=terms}).
The data and code are released with an MIT license.
}

\dsquestionex{Have any third parties imposed IP-based or other restrictions on the data associated with the instances?}{If so, please describe these restrictions, and provide a link or other access point to, or otherwise reproduce, any relevant licensing terms, as well as any fees associated with these restrictions.}
\label{Q49}

\dsanswer{
The videos corresponding to our instances are legally owned by YouTube users.
Our dataset users can download them from the URLs we provide in annotation files, but redistributing videos for commercial use is prohibited.
}

\dsquestionex{Do any export controls or other regulatory restrictions apply to the dataset or to individual instances?}{If so, please describe these restrictions, and provide a link or other access point to, or otherwise reproduce, any supporting documentation.}
\label{Q50}

\dsanswer{No.}

\dsquestion{Any other comments?}
\label{Q51}

\dsanswer{No.}

\end{compactenum}

\section*{Maintenance}

\begin{compactenum}[\hspace{0pt}Q1.]
\setcounter{enumi}{51}

\dsquestion{Who will be supporting/hosting/maintaining the dataset?}
\label{Q52}

\dsanswer{
The authors will maintain the dataset.
The dataset is hosted using Inria servers and Google Drive service.
All the information about the dataset, including links to the paper, code, and future announcements will be accessible at the dataset website~\cite{chapterswebpage}.
}

\dsquestion{How can the owner/curator/manager of the dataset be contacted (e.g., email address)?}
\label{Q53}

\dsanswer{The contact emails of authors are available on the dataset website~\cite{chapterswebpage}.}

\dsquestionex{Is there an erratum?}{If so, please provide a link or other access point.}
\label{Q54}

\dsanswer{
There is no erratum for our initial release.
We will version all errata as future releases (\qref{Q55}) and document them on the dataset website~\cite{chapterswebpage}.
}

\dsquestionex{Will the dataset be updated (e.g., to correct labeling errors, add new instances, delete instances)?}{If so, please describe how often, by whom, and how updates will be communicated to users (e.g., mailing list, GitHub)?}
\label{Q55}

\dsanswer{
We will update our dataset once every year and announce it on the dataset website~\cite{chapterswebpage}.
These future versions would remove instances that were requested to be removed via the opt-out form (\qref{Q32}).
}

\dsquestionex{If the dataset relates to people, are there applicable limits on the retention of the data associated with the instances (e.g., were individuals in question told that their data would be retained for a fixed period of time and then deleted)?}{If so, please describe these limits and explain how they will be enforced.}
\label{Q56}

\dsanswer{
Rather than directly distributing videos, we distribute URLs that point to the original videos uploaded by YouTube users.
This means that users retain full control of their data -- any post deleted from YouTube will be automatically removed from \dataset{} (see also \qref{Q10}, \qref{Q14}, \qref{Q31}).
}

\dsquestionex{Will older versions of the dataset continue to be supported/hosted/maintained?}{If so, please describe how. If not, please describe how its obsolescence will be communicated to users.}
\label{Q57}

\dsanswer{
A new version release of \dataset{} will automatically deprecate its previous version.
We will only support and maintain the latest version at all times.
Deprecated versions will remain accessible on the dataset website for a few weeks, after which they will be removed.
We decided to deprecate old versions to ensure that any data that is requested to be removed (\qref{Q32}) will be no longer accessible in future versions.
}

\dsquestionex{If others want to extend/augment/build on/contribute to the dataset, is there a mechanism for them to do so?}{If so, please provide a description. Will these contributions be verified? If so, please describe how. If not, why not? Is there a process for communicating/distributing these contributions to other users? If so, please provide a description.}
\label{Q58}

\dsanswer{
Anyone can extend \dataset{} by using our data collection code (linked on our website~\cite{chapterswebpage}).
We are open to accepting extensions via personal communication with contributors.
Otherwise, our code and data licenses allow others to create independent derivative works (with proper attribution) as long as they are used for non-commercial academic research.
}
\end{compactenum}

\end{document}